\documentclass[sensors,article,accept,pdftex,moreauthors]{Definitions/mdpi} 
\firstpage{1} 
\pubvolume{25}
\issuenum{1}
\articlenumber{3475}
\pubyear{2025}
\copyrightyear{2025}
\externaleditor{Pavlos Lazaridis, Christos Tachtatzis and Euler Cássio Tavares De Macêdo} % More than 1 editor, please add `` and '' before the last editor name
\datereceived{1 May 2025} 
\daterevised{26 May 2025} % Comment out if no revised date
\dateaccepted{28 May 2025} 
\datepublished{31 May 2025} 
\hreflink{https://doi.org/10.3390/s25113475} % If needed use \linebreak
\doinum{10.3390/s25113475}
\pdfoutput=1 % Uncommented for upload to arXiv.org
\usepackage{enumerate}
\usepackage{subfiles}
\usepackage[nottoc]{tocbibind}
\usepackage{lipsum}
\usepackage{mathtools}
\usepackage[noend]{algpseudocode}
\usepackage{enumitem,kantlipsum}
\usepackage{indentfirst}
\usepackage{multirow}
\usepackage{pifont}
\usepackage{amsmath,amssymb,amsfonts}
\usepackage{algorithm}
\usepackage{graphicx}
\usepackage{textcomp}
\usepackage{bbding}
\usepackage{rotating}
\usepackage{subcaption}

\usepackage[bottom,flushmargin]{footmisc}
\usepackage{adjustbox}

\newcommand\footnoteref[1]{\protected@xdef\@thefnmark{\ref{#1}}\@footnotemark}
\usepackage{multicol,siunitx,ragged2e}
\newlength\mylen
\newcolumntype{K}[1]{>{\centering\arraybackslash}p{#1}}
\Title{Data Quality Monitoring for the Hadron Calorimeters Using Transfer Learning for Anomaly Detection }

\TitleCitation{Data Quality Monitoring for the Hadron Calorimeters Using Transfer Learning for Anomaly Detection}

\Author{{Mulugeta Weldezgina Asres} $^{1,}$*\orcidA{}, Christian Walter Omlin $^{1,}$*\orcidB{}, {Long Wang} $^{2}$\orcidC{}, {David Yu} $^{3}$\orcidD{}, {Pavel Parygin} $^{4}$\orcidE{}, {Jay Dittmann} {$^{5}$\orcidF{}} 
 and {the CMS-HCAL Collaboration} {$^{\dagger}$} %MDPI: We revised the symbol, please confirm
}

\AuthorNames{Mulugeta Weldezgina Asres, Christian Walter Omlin, Long Wang, David Yu, Pavel Parygin, Jay Dittmann, and the CMS-HCAL Collaboration.
}

\AuthorCitation{Asres, M.W.; Omlin, C.W.; Wang, L.; Yu, D.; Parygin, P.; Dittmann, J.; the CMS-HCAL Collaboration
}
\address{$^{1}$ \quad {Centre for Artificial Intelligence Research, Department of Information and Communication Technology}, {University~of~Agder}, {4879 Grimstad}, {Norway} 
 \\
$^{2}$ \quad {Department of Physics}, {University~of~Maryland}, {College~Park, MD~20742}, USA; long.wang@cern.ch\\
$^{3}$ \quad {Department of Physics}, {Brown~University}, {Providence, RI~02912}, USA; david\_yu@brown.edu\\
$^{4}$ \quad {Department of Physics and Astronomy}, {University~of~Rochester}, {Rochester, NY~14627}, USA; pavel.parygin@cern.ch\\
$^{5}$ \quad {Department of Physics}, {Baylor~University}, {Waco, TX~76706}, USA; jay\_dittmann@baylor.edu\\
}

\corres{Correspondence: mulugetawa@uia.no (M.W.A.); christian.omlin@uia.no (C.W.O.)}
\firstnote{The CMS-HCAL Collaboration author list is given at {Supplementary Materials}}
\abstract{
The proliferation of sensors brings an immense volume of spatio-temporal (ST) data in many domains, including monitoring, diagnostics, and prognostics applications. Data curation is a time-consuming process for a large volume of data, making it challenging and expensive to deploy data analytics platforms in new environments. Transfer learning (TL) mechanisms promise to mitigate data sparsity and model complexity by utilizing pre-trained models for a new task. 
Despite the triumph of TL in fields like computer vision and natural language processing, efforts on complex ST models for anomaly detection (AD) applications are limited. 
In this study, we present the potential of TL within the context of high-dimensional ST AD with a hybrid autoencoder architecture, incorporating convolutional, graph, and recurrent neural networks. Motivated by the need for improved model accuracy and robustness, particularly in scenarios with limited training data on systems with thousands of sensors, this research investigates the transferability of models trained on different sections of the Hadron Calorimeter of the Compact Muon Solenoid experiment at CERN. The key contributions of the study include exploring TL's potential and limitations within the context of encoder and decoder networks, revealing insights into model initialization and training configurations that enhance performance while substantially reducing trainable parameters and mitigating data contamination effects.
Code: \href{https://github.com/muleina/CMS_HCAL_ML_OnlineDQM}{https://github.com/muleina/CMS\_HCAL\_ML\_OnlineDQM}
}


\begin{document}
%\begin{sloppypar}

\section{Introduction}
\label{sec:introduction}

Spatio-temporal (ST) anomaly detection (AD) a promising monitoring application of deep learning~(DL) in several fields~\cite{atluri2018spatio, chang2022video, deng2022graph, 
tivsljaric2021spatiotemporal, fathizadan2023deep, mulugeta2022dqm}.
A unique quality of ST data is the presence of dependencies among measurements induced by the spatial and temporal attributes, where data correlations are more complex to capture using conventional techniques~\cite{atluri2018spatio}. 
A spatio-temporal anomaly can thus be defined as a data point or cluster of data points that violate the nominal ST correlation structure of the normal data points.
DL models dominate the recent AD studies, as~AD models capture complex structures, extract end-to-end automatic features, and~scale for large-volume data sets~\cite{zhao2022comparative, mulugeta2022dqm, chalapathy2019deep, cook2019anomaly, fathizadan2023deep, wang2024self, yang2025weighted}. AD models can broadly be categorized as: (1) supervised methods requiring labeled anomaly observations~\cite{wang2024self, yang2025weighted} and~(2) unsupervised approaches using unlabeled data, which are more pragmatic in many real-world applications, as~data labeling is tedious and expensive~\cite{zhao2022comparative, mulugeta2022dqm, chalapathy2019deep, cook2019anomaly, fathizadan2023deep}. Unsupervised AD models trained with only healthy observations are often called semi-supervised approaches~\cite{mulugeta2022dqm}.
Semi-supervised AD models have accomplished promising performance in reliability, safety, and~health monitoring applications in several domains~\cite{mulugeta2022dqm, chalapathy2019deep, zhao2022comparative, cook2019anomaly}. 

The deployment of ST~DL models in a new environment is often circumscribed by the limited amount of clean data~\cite{wang2021spatio}.
Data curation for DL modeling remains cumbersome and particularly challenging for temporal data despite abundant availability. 
Transfer learning~(TL) mechanisms have been proposed for DL models to mitigate the challenge of data insufficiency; it accelerates model training and enhances accuracy~\cite{yu2022survey, shao2018highly, laptev2018reconstruction, gupta2018transfer, boulle2020classification, wang2021spatio, wang2018cross, hijazi2023transfer, shao2014transfer, niu2020decade, wang2024self, yang2025weighted,  adama2021survey, chato2023survey}. 
It aims to achieve in-domain and cross-domain learning by extracting useful information from the model or data of the source task and transferring it to the target tasks~\cite{shao2014transfer, yu2022survey, niu2020decade,  adama2021survey, chato2023survey}. 
TL is widely employed in computer vision (e.g., a~large image classifier trained on over 1000~classes with \textsc{ImageNet1K}~\cite{ILSVRC15} is fine-tuned to classify a few types of fruit categories)~\cite{shao2014transfer} and~natural language processing (e.g., BERT~\cite{devlin2018bert}, initially trained on a massive and diverse text corpus to learn general language features like syntax and semantics, is fine-tuned with smaller task-specific data sets, specialized in question-answering tasks)~\cite{niu2020decade}. It has also been proposed for temporal sensor data related to machine monitoring~\cite{shao2018highly}, electricity load monitoring~\cite{laptev2018reconstruction}, medical applications~\cite{gupta2018transfer}, dynamic systems~\cite{boulle2020classification}, and~ST data for crowd prediction~\cite{wang2018cross, wang2021spatio, sarker2021semi, natha2025deep}, finance~\cite{guo2018citytransfer}, environment monitoring~\cite{sarker2021semi}, and fault diagnosis~\cite{hijazi2023transfer, yang2025weighted}. TL on ST data for AD application remains limited, and~deeper investigation on autoencoder (AE) models, assessing both the encoder and decoder networks, is lacking~\cite{wang2018cross, wang2021spatio, guo2018citytransfer, hijazi2023transfer, natha2025deep, yang2025weighted}.

Our study discusses ST AD modeling for the \textit{{Compact} %MDPI: Please check all the italics in the whole paper and confirm if the italics are unnecessary and can be removed. If yes, please revise all of them.
% The italics are necessary for emphasising the terms. Hence, they should be kept!
 Muon Solenoid}~(CMS) experiment at the \textit{Large Hadron Collider}~(LHC)~\cite{evans2008lhc, cms2023development}. 
The CMS experiment, one of the two high-luminosity general-purpose detectors at the LHC, consists of a tracker to reconstruct particle paths accurately, two calorimeters---the \textit{electromagnetic}~(ECAL) and the \textit{hadronic}~(HCAL)---to detect electrons, photons, and hadrons, and~a \textit{muon} system~\cite{collaboration2008cms, cms2023development}.
The CMS experiment employs the \textit{Data Quality Monitoring}~(DQM) system to guarantee high-quality physics data through online monitoring that provides live feedback during data acquisition, as well as~offline monitoring that certifies the data quality after offline processing~\cite{azzolini2019data}. The~online DQM identifies emerging problems using reference distributions and predefined tests to detect known failure modes using summary histograms, such as digi-occupancy maps of the calorimeters~\cite{tuura2010cms, de2014cms}. 
A digi-occupancy map contains the histogram record of particle hits of the data-taking channels of the calorimeters at the digitization level. 
The CMS calorimeters may encounter problems during data taking, such as issues with the front-end particle sensing scintillators, digitization and communication systems, back-end hardware, and~algorithms~\cite{azzolin2019improving, mulugeta2022dqm}. These problems are usually reflected in the digi-occupancy maps. 
The growing complexity of detectors and the variety of physics experiments make data-driven AD systems essential tools for CMS to automate the detection, identification, and~localization of detector anomalies~\cite{mulugeta2022dqm, asres2021unsupervised, asres2022long, cms2023autoencoder}. 
Recent efforts in DQM at CMS have presented DL for AD applications~\cite{azzolin2019improving, azzolini2019data, pol2019anomaly, pol2019detector, mulugeta2022dqm, cms2023autoencoder, parra2024human}.
The synergy in AD has thus far achieved promising results on spatial 2D histogram maps of the DQM for the ECAL~\cite{azzolin2019improving, cms2023autoencoder}, the~muon detectors~\cite{pol2019detector}, and ST 3D maps of the HCAL~\cite{mulugeta2022dqm}. 

Further study of TL for ST AD models---often involving combinations of spatial and temporal learning networks---is essential, considering the achievements of TL in other domains~\cite{niu2020decade}.
Recent ST DL models are hybrid and commonly made of combinations of variants of convolutional neural networks~(CNNs), recurrent neural networks~(RNNs), graph neural networks~(GNNs), and~transformers for various data mining tasks~\cite{xie2020urban, wang2018cross, wang2021spatio, lai2025stglr, natha2025deep}. 
Our study investigates the potential strengths and limitations of TL on high-dimensional ST semi-supervised AD models. 
Although there are several ST AD architectures in the literature, most operate with 2D spatial data, such as images~\cite{sarker2021semi, wang2018cross, wang2021spatio, wang2024self, natha2025deep}, and~the ones that incorporate GNN deal with a limited number of nodes ranges from tens to a few hundreds~\cite{lai2025stglr,deng2022graph}. 
Hence, we limit our discussion to the \textsc{GraphSTAD} system~\cite{mulugeta2022dqm}---an AE model made of a CNN, RNN, and~GNN operating on high-dimensional 3D spatial data---to deeply investigate the potential of TL for ST AD in the context of calorimeters for the CMS experiment. The~\textsc{GraphSTAD} has been proposed for online DQM to automate monitoring of the thousands of HCAL channels through DL in Ref.~\cite{mulugeta2022dqm}. The~model captures abnormal events using spatial appearance and temporal context on digi-occupancy maps of the DQM. 
\textsc{GraphSTAD} employs CNNs to capture the behavior of adjacent channels exposed to regional collision particle hits, GNNs to learn local electrical and environmental characteristics due to a shared back-end circuit of the channels, and~RNNs to detect temporal degradation on faulty channels~\cite{mulugeta2022dqm}. 
We have transferred a pre-trained \textsc{GraphSTAD} model on the source \textit{HCAL Endcap} (HE) subsystem into another target subsystem of the \textit{HCAL Barrel} (HB) for the TL experiment. The~HE and HB are subdetectors of the HCAL; they are designed to capture hadron particles at different positions of the calorimeter. The~subdetectors share similarities but also have differences in design, technology, and~configuration, such as detector segmentation~\cite{strobbe2017upgrade}.

Brute-forcing the knowledge from the source into the target, irrespective of their divergence, and thorough investigation of the several network-building modules would cause certain performance degeneration~\cite{shao2014transfer, adama2021survey}.
Hence, we have provided insights on TL using various training modes with different network hierarchies of the AE of the \textsc{GraphSTAD} system. The~experiment has demonstrated the potential of TL when applied to the feature extraction encoder and the reconstruction decoder networks with different fine-tuning mechanisms on the target dataset. 
We have also examined the impact of TL with RNN state preservation within and across sliding time windows on ST reconstruction. 
TL has achieved promising ST reconstruction and AD while reducing the trainable parameters and providing better robustness against anomaly contamination in the training dataset. 
Our study demonstrates the efficacy of ST TL in overcoming training data sparsity and model training~computation. 

The key contributions of this study can be summarized~as follows:
\begin{itemize}
    \item This study explores the potential and limitations of TL in the context of high-dimensional ST models for AD application, at~scale with 3000--7000 spatial nodes.
    \item This study, different from existing TL studies, assesses both encoder and decoder networks of a hybrid AE---evaluated on each main building block with various configurations. Related TL studies primarily focus on the feature extraction encoder or fine-tuning the entire network, as~highlighted in this study. We present deeper insights and considerations on previously underexplored angles of TL. 
    \item We demonstrate the robustness of TL, with~limited training data sets, in~improving model accuracy, reducing training parameters, and~better mitigation against vulnerability to training data contamination.
\end{itemize}

We discuss the related work in TL and the CMS DQM system in Section~\ref{sec:background}. We describe our datasets in Section~\ref{sec:datasetdescription} and the AD and TL methodologies in Section~\ref{sec:methodology}. Section~\ref{sec:resultsanddiscussion} presents the performance evaluation and discussion of the results. We provide the conclusion and review the impact of our results in Section~\ref{sec:conclusion}.

\section{Background}
\label{sec:background}

This section discusses TL in DL models and provides an overview of the DQM system of the CMS~experiment.

\subsection{Transfer Learning on Deep~Learning}

In the last decade, the~effectiveness of DL in handling large datasets has caught the attention of both academia and industry. Its ability to learn nonlinear behavior, along with end-to-end automatic feature extraction, allows it to find complex patterns within high-dimensional large data sets. However, most DL models are complex and require extensive data sizes for modeling, which can be expensive and time-consuming to curate, especially in the case of temporal data. 
Transfer learning approaches, which incorporate pre-trained models into new tasks, are potential solutions for developing DL models when clean data are limited~\cite{shao2018highly, laptev2018reconstruction, gupta2018transfer, boulle2020classification, hijazi2023transfer, huber2024leveraging}.
{TL is a paradigm where knowledge from a source model or data on different domains (e.g., different data sources or datasets) or tasks (e.g., different model applications) is utilized to improve the efficacy of a target model~\cite{yu2022survey,  adama2021survey, chato2023survey, shao2014transfer, huber2024leveraging}.}

{The TL techniques in the literature can broadly be categorized into various taxonomies~\cite{shao2014transfer, niu2020decade, yu2022survey}. One of the typical categorizations is based on the similarity of the task and domain between the source and target~\cite{shao2014transfer, yu2022survey, adama2021survey, niu2020decade}: (1) inductive TL: the source and target tasks are different, but their domains may remain the same; (2) transductive TL: the tasks remain the same, but~the domains are different; and~(3) unsupervised TL: similar to inductive transferring on different but related tasks with unlabeled datasets. TL can be carried out on (1) model parameters, where all or some parameters are transferred from a pre-trained source model, and~(2) data, where all or part of the source domain data instances are utilized to train the target model~\cite{niu2020decade, yu2022survey, adama2021survey}. In~this study, TL signifies the use of learned network parameters (\textit{weights} and \textit{biases} from a source model pre-trained on adequate datasets) on a target model for a related task on a different dataset, with~or without fine-tuning of the parameters~\cite{yu2022survey}. We refer readers to a survey study in refs.~\cite{niu2020decade, yu2022survey, adama2021survey} for further discussion and progress on recent deep TL approaches.}

The recent successes of generative models on image and text data have ameliorated the adoption of TL methods for several applications~\cite{shao2014transfer, niu2020decade, yu2022survey}. 
The notable contribution of TL is significant in transferring feature extraction networks (\textit{encoders}) that are trained on immense datasets with very expensive computation grids~\cite{sarker2021semi, natha2025deep}. Robustly extracted features reduce the model complexity and training cost of the fine-tuned decision networks while enhancing accuracy~\cite{gupta2018transfer, shao2018highly, niu2020decade, wang2018cross, sarker2021semi, natha2025deep}. 
We refer to this TL mechanism as the {freeze and fine-tune} approach~\cite{yu2022survey}.
Although abundant studies are available for images and language text, TL is relatively less explored for temporal data, such as sensor measurement datasets~\cite{hijazi2023transfer}; TS datasets are often not readily available or accessible on the internet, unlike images and text, and~the datasets are often multidimensional and so diverse that they require domain-specific knowledge for data curation and preparation.  
TL on a temporal data has been investigated in various applications~\cite{shao2018highly, laptev2018reconstruction, gupta2018transfer, boulle2020classification, yang2025weighted}.
The efforts towards adopting TL for ST data are even more limited~\cite{wang2018cross, wang2021spatio, guo2018citytransfer, hijazi2023transfer, sarker2021semi, natha2025deep}. 
Hijazi~et~al.~\cite{hijazi2023transfer} proposed a TL approach that integrates CNN and temporal long short-term memory (LSTM) (referred to as \textsc{ConvLSTM}) to efficiently train a new stability prediction model when the power system undergoes topological changes. 
Wang~et~al.~\cite{wang2018cross} applied TL for cross-city crowd-flow prediction where feature extraction \textsc{ConvLSTM} of the forecasting model trained on one city is fine-tuned on another city's dataset. 
Wang~et~al.~\cite{wang2021spatio} extended TL on \textsc{ConvLSTM} using a deep adaptation mechanism for crowd-flow prediction. The~adaptation network matches the embedding representations of the source and target domain distributions to learn the transferable features between the two domains.
Guo~et~al.~\cite{guo2018citytransfer} fine-tuned an autoencoder for a store recommendation system from a model trained on a different city dataset. 
Sarker~et~al.~\cite{sarker2021semi} and Natha~et~al.~\cite{natha2025deep} adopted pretrained 3D CNNs for ST feature extraction to improve anomaly detection on video datasets.
Yang~et~al.~\cite{yang2025weighted} proposed unsupervised TL utilizing the fault knowledge learned from labeled sensor fault datasets to perform online anomaly monitoring on unmanned aerial vehicle sensor data.
Some studies have employed TL to increase training data from multiple sources and address training with diverse data issues, such as catastrophic forgetting, using different regularization techniques~\cite{huber2024leveraging}. 
Recent DL models built on hybrids of \mbox{CNNs~\cite{chang2022video, wu2020fast, hasan2016learning, hijazi2023transfer, wang2018cross, mulugeta2022dqm, sarker2021semi, natha2025deep}}, \mbox{RNNs~\cite{luo2019video, hsu2017anomaly, hijazi2023transfer, wang2018cross, mulugeta2022dqm, natha2025deep, huber2024leveraging}}, and \mbox{GNNs~\cite{hsu2017anomaly, deng2022graph, mulugeta2022dqm, lai2025stglr}} have gained momentum for TS and ST data in AD and other data mining applications. Thus far, most TL studies have focused on feature extraction encoding networks and predominantly on forecasting tasks~\cite{wang2018cross, wang2021spatio, hijazi2023transfer, huber2024leveraging}.  
We have studied the transferability of CNNs, GNNs, and~RNNs on both the encoder and decoder networks of an autoencoder and qualitatively evaluated the effectiveness of the TL on ST reconstruction and AD~tasks.

\subsection{The Hadron Calorimeter of the CMS~Detector}

\begin{figure}[H]
\subfloat[\centering]{\includegraphics[width=0.68\textwidth]{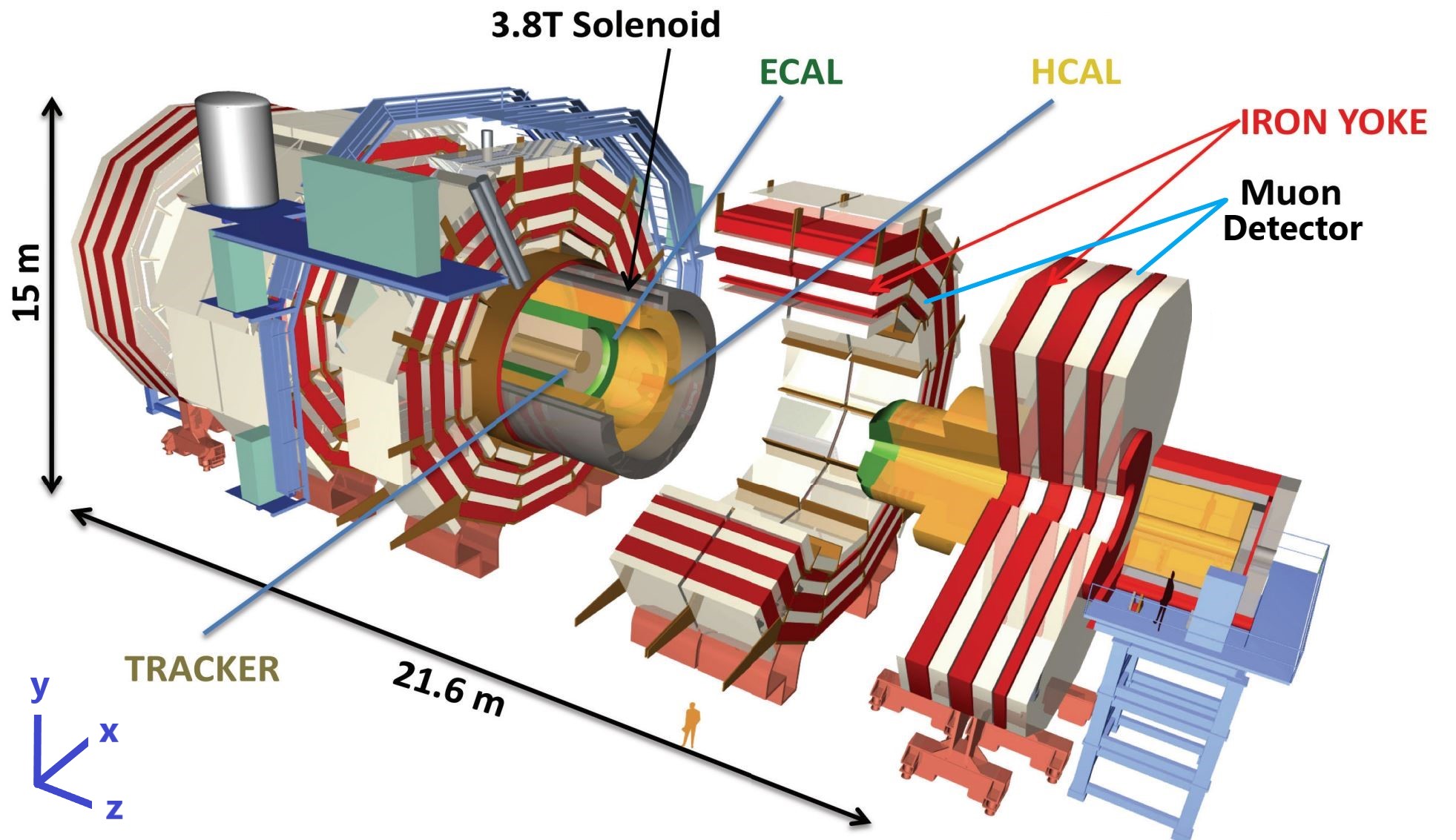}}
%\hfill
\subfloat[\centering]{\includegraphics[width=0.3\textwidth]{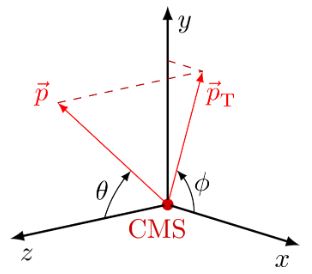}}
%\centering
%\captionsetup[subfigure]{justification=centering}
%\begin{subfigure}[]{0.68\textwidth}
%\centering
%\includegraphics[width=1\textwidth, scale=1]{images/Schematic_view_of_the_CMS.jpg}
%\caption{}
% \medskip
%\label{fig:cms_diagram}
%\end{subfigure}
%\hfill
%\centering
%\begin{subfigure}[]{0.3\textwidth}
%\centering
%\includegraphics[width=1\textwidth, scale=1]{images/cms_eta_phi_coordinates.jpg}
%\caption{}
% \medskip
%\label{fig:cms_diagram_eta_phi}
%\end{subfigure}
\caption{{Schematic} %MDPI: 1. We adjusted the positions of the subfigures, please confirm. 2. Please note that the size of figures will be adjusted appropriately to ensure the clarity and readability of the image. Changes to the position of figures and tables may occur during the production stage.
 of the CMS detector: (\textbf{a}) CMS with its major systems~\cite{focardi2012status}, and~(\textbf{b}) geometry axes and angles of the CMS with respect to the collision intersection point~\cite{tikz2023}.\label{fig:cms_diagram_and_eta_phi}}
\end{figure}

Figure~\ref{fig:cms_diagram_and_eta_phi}a shows the CMS experiment and the HCAL detector inside CMS~\cite{collaboration2008cms, cms2023development}. 
The calorimeters of the CMS detector are highly segmented to improve the accuracy of energy-deposition profile-measurement and particle identification~\cite{collaboration2008cms, cms2023development, focardi2012status}. 
The segmentation geometry of the detector is represented using $\eta$ and $\phi$ spaces, which correspond to \textit{pseudo-rapidity} and \textit{azimuth}, respectively (as shown in Figure~\ref{fig:cms_diagram_and_eta_phi}b). 
The $z$-axis lies along the incident beam direction, $\phi$ is the azimuthal angle between the $x$ and $y$ axis, and~$\eta$ is calculated from the polar angle $\theta _ {cm}$ between the $z$ and $xy-$planes as {follows:} %MDPI: Please carefully check variable formatting (italic, bold, subscript, uppercase, etc.) throughout the manuscript to ensure the formatting is consistent and revise if needed, and please confirm if there is duplicated equations, if yes, please confirm if they are right
%Author: we confirm the formatting is correct and should be kept as such.
\begin{linenomath}\begin{equation}
    \eta = -\ln (\tan(\theta _ {cm}/2))
\end{equation}\end{linenomath}
where $x$, $y$, and~$z$ are orthogonal axes of the cylinder,~$\theta _ {cm}$ is the center-of-mass scattering angle, and $\ln$ is a natural log function. 
The $\eta - \phi$ space corresponds to a rectangular coordinate system representing an outgoing particle's direction from the center of the detector (where the collision occurs). 
Particles traveling in the same direction lie near each other in $\eta - \phi$ space. 

\begin{figure}[H]

%\captionsetup[subfigure]{justification=centering}
\begin{adjustwidth}{-\extralength}{0cm}\centering
\subfloat[\centering]{\includegraphics[width=0.65\textwidth]{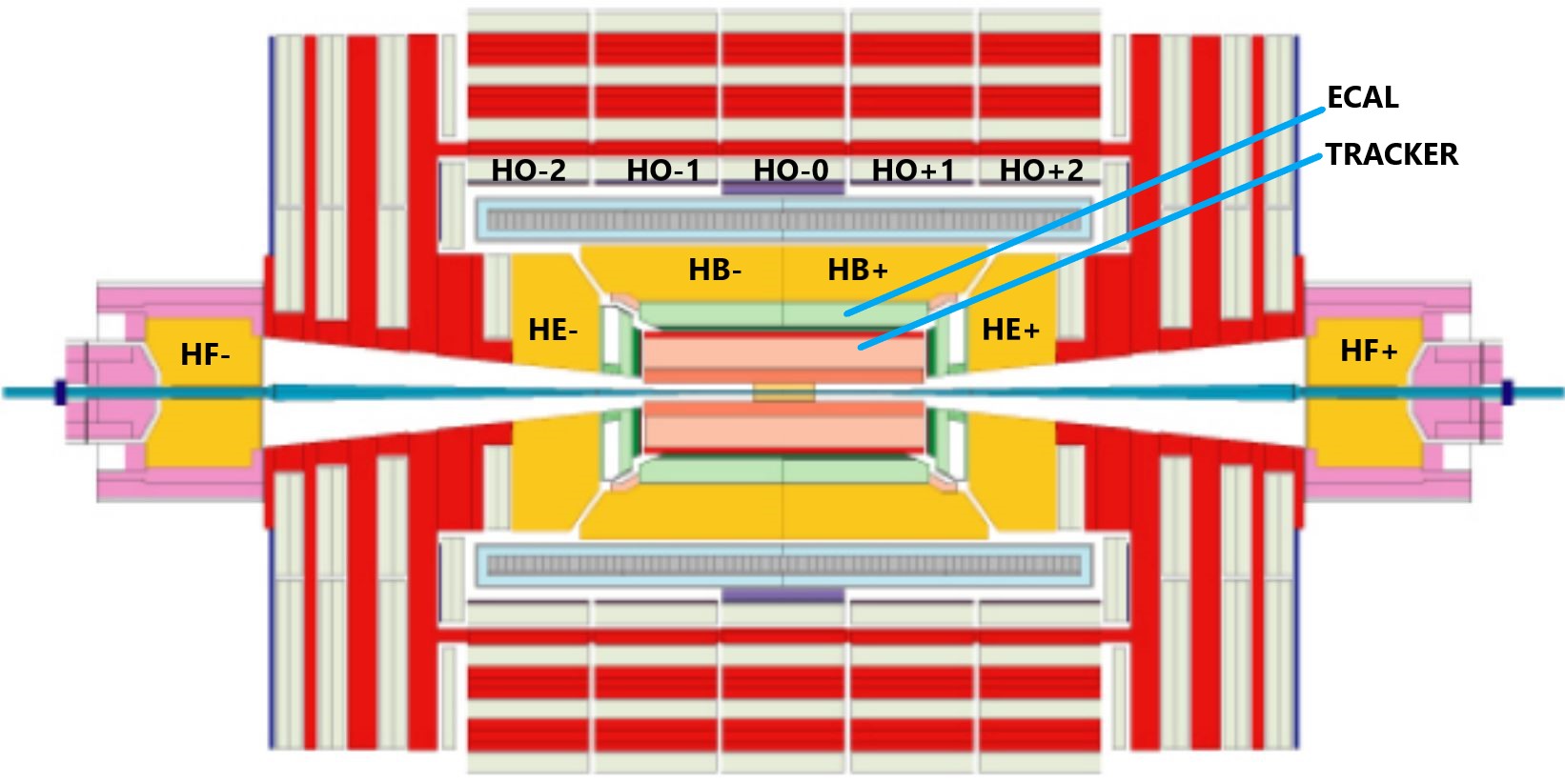}}
% \medskip
\subfloat[\centering]{\includegraphics[width=0.65\textwidth]{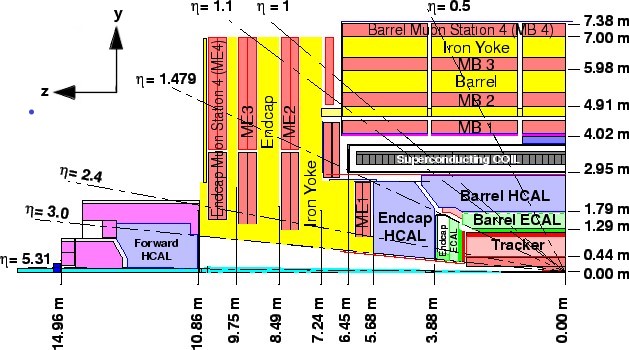}}
% \medskip
\end{adjustwidth}
\caption{{The} %MDPI: The contents of this figure are not clear/overlapping. Please confirm if affect reading, if yes, please replace the image with one of a sufficiently high resolution and ensure all the text are legible.
%The most relevant content of the figures are legible and should be kept as such.
 subdetectors of the HCAL: (\textbf{a}) longitudinal view of the HB, HE, HF, and~HO subdetectors on CMS~\cite{cheung2012cms}; and (\textbf{b}) longitudinal view of one quadrant of CMS with segmentation angle specifications of the $\eta$, where the origin denotes the interaction point~\cite{collaboration1999cms, collaboration2008cms}.\label{fig:cms_diagram_hcal}}
\end{figure}

Figure~\ref{fig:cms_diagram_hcal}a illustrates the four major subdetectors of the HCAL covering different segments in the CMS detector: the HB, the~HE, the~\textit{HCAL Outer} (HO), and~the \textit{HCAL Forward} (HF). 
Since this study's datasets are from the LHC Run-2 collision experiment, we will describe the HCAL system configurations from 2018 below. 
The HB and HE are sampling calorimeters with a brass absorber and active plastic scintillators to measure the energy depositions~\cite{collaboration2008cms}. 
The subdetectors surround the ECAL and are fully immersed within the strong magnetic field of the solenoid: the HB are joined hermetically with the barrel extending out to $\left| \eta \right|=1.4$ and~the HE covering the overlapping range $1.3<\left| \eta \right|<3.0$ (as shown in Figure~\ref{fig:cms_diagram_hcal}b). The~HF is located 11.2 m from the interaction point and extends the pseudo-rapidity coverage (overlapping with the HE) from $\left| \eta \right|=2.9$ to $\left| \eta \right|=5$.  
The central shower containment in the region $\left| \eta \right|<1.26$ is improved with the HO, an~array of scintillators located outside the~magnet. 

\vspace{-9pt}

The {front-end electronics} of the HCAL, responsible for sensing and digitizing optical signals of the collision particles, are divided into sectors of \textit{readout boxes} (RBXes) that house the electronics and provide voltage, backplane communications, and~cooling.
Our study's use cases, the~HE and HB, consist of 36 RBXes arranged on the plus (HE[HB]P) and minus (HE[HB]M) hemispheres of the CMS detector. 
The front-end acquisition systems transmit the photons produced in the plastic scintillators through the wavelength-shifting fibers to the {silicon photomultipliers} (SiPMs) [HE] or the {hybrid photodiode transducers} (HPD) [HB]~\cite{collaboration2008cms}.
Each RBX houses frontend electronics that include four digitization {readout modules} (RMs), the~{next-generation clock and control module}, and~the {calibration unit}~\cite{collaboration2008cms}.
Each RM is made of SiPMs [HE] or HPD [HB], a~SiPM control card, and~four readout {charge integrator and encoder} (QIE) cards, each with several QIE chips and {field-programmable gate array} (FPGA) modules. 
A QIE chip integrates charge from one SiPM [HE] or HPD [HB] at 40 MHz, and~the FPGA serializes and encodes the data from the QIE chips (channels).

\subsection{CMS Data Quality~Monitoring}
\label{sec:hcal_dqm_rbx}

The collision data of the LHC are organized into {runs}, where each run contains thousands of luminosity sections (also called lumisections). A {lumisection} (LS) corresponds to approximately 23 s of data collection and comprises hundreds or thousands of collision events containing particle hit records across the CMS detector.
The DQM system in CMS provides feedback on detector performance and data reconstruction; it generates a list of certified runs for physics analyses and stores it in the ``Golden JSON''~\cite{azzolini2019data}. 
The DQM employs online and offline monitoring mechanisms: (1) {online monitoring} is real-time DQM during data acquisition, and~(2) {offline monitoring} provides the final fine-grained data quality analysis for data certification 48 h after the collisions were recorded. 
The online DQM populates a set of histogram-based maps on a selection of events and provides summary plots with alarms that DQM experts inspect to spot problems. 
The {digi-occupancy} map is one of the histogram maps generated by the online DQM, and~it contains particle hit histogram records of the particle readout channel sensor of the calorimeters. 
A digi, also called a hit, is a reconstructed and calibrated collision physics signal of the calorimeter. 
Several errors can arise in the calorimeter affecting the front-end particle sensing scintillators, the~digitization and communication systems, the~back-end hardware, or~the algorithms. These errors appear in the digi-occupancy map as holes, under- or over-populated bins, or~saturated bins. 
Previous efforts by the authors of refs.~\cite{azzolin2019improving, azzolini2019data, pol2019anomaly, pol2019detector, mulugeta2022dqm} demonstrate the promising AD efficacy of using digi-occupancy maps for calorimeter channel monitoring using machine learning.
Our \textsc{GraphSTAD} has extended the efforts in AD for the HCAL with ST modeling of the 3D digi-occupancy maps of the DQM~\cite{mulugeta2022dqm}. 
The \textsc{GraphSTAD} incorporates both CNNs and GNNs to capture Euclidean and non-Euclidean spatial characteristics, respectively, as well as~RNNs for temporal learning for the HCAL~channels.

\section{Dataset~Description}
\label{sec:datasetdescription}

\begin{figure}[H]

%\captionsetup[subfigure]{justification=centering}
\begin{adjustwidth}{-\extralength}{0cm}\centering
\subfloat[\centering]{\includegraphics[width=0.445\textwidth]{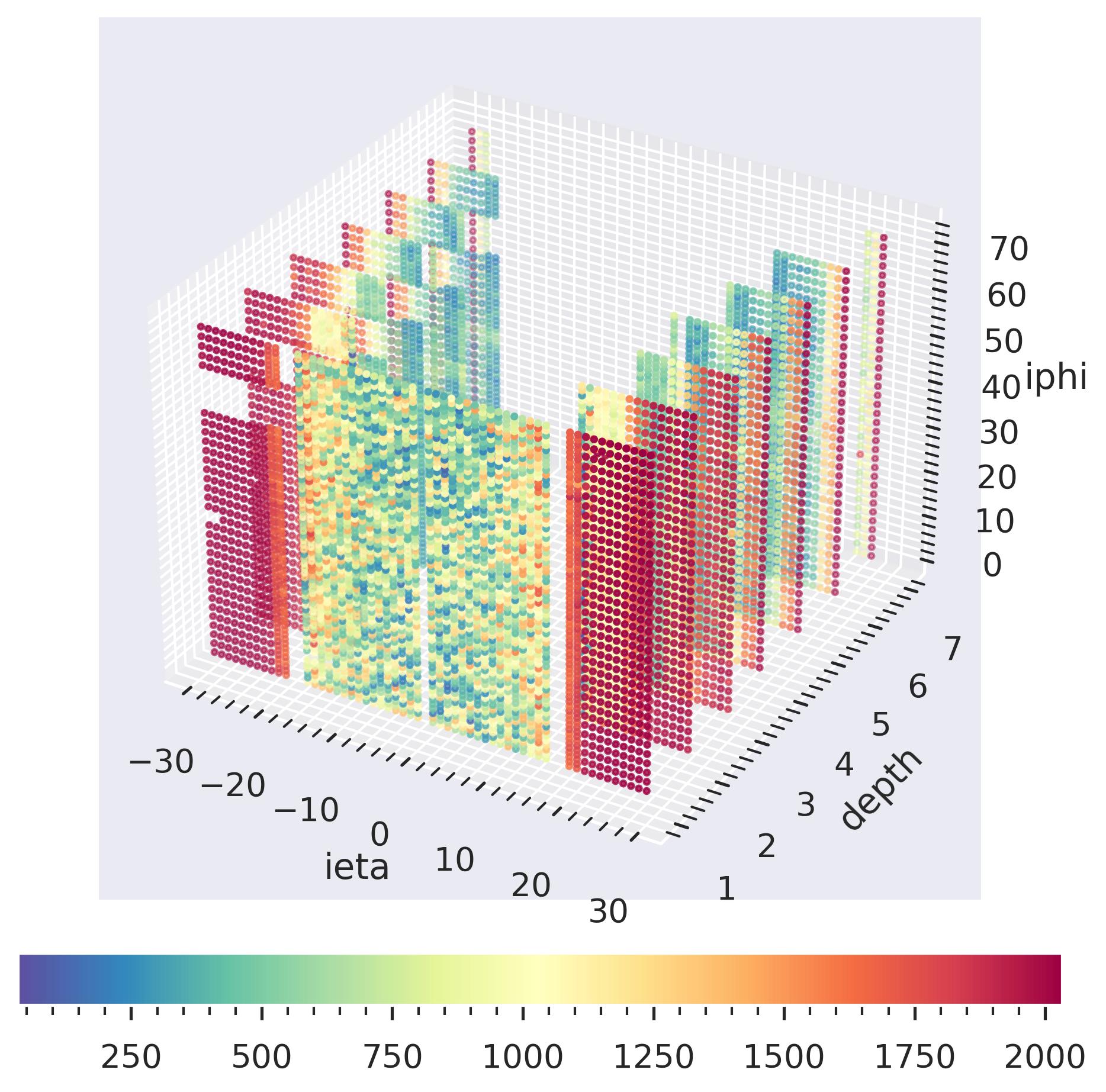}}
% \medskip
\subfloat[\centering]{\includegraphics[width=0.445\textwidth]{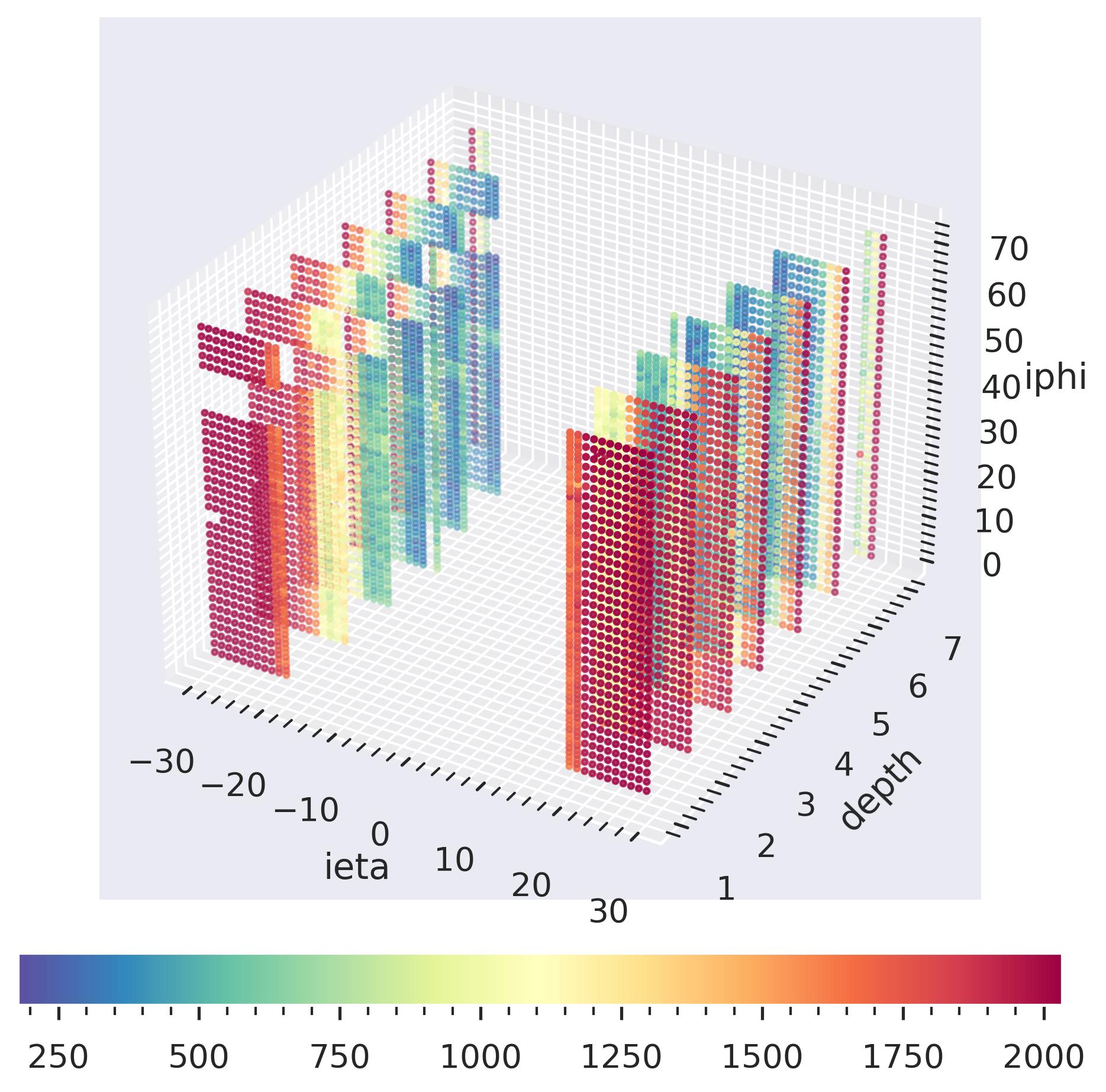}}
% \medskip
\subfloat[\centering]{\includegraphics[width=0.445\textwidth]{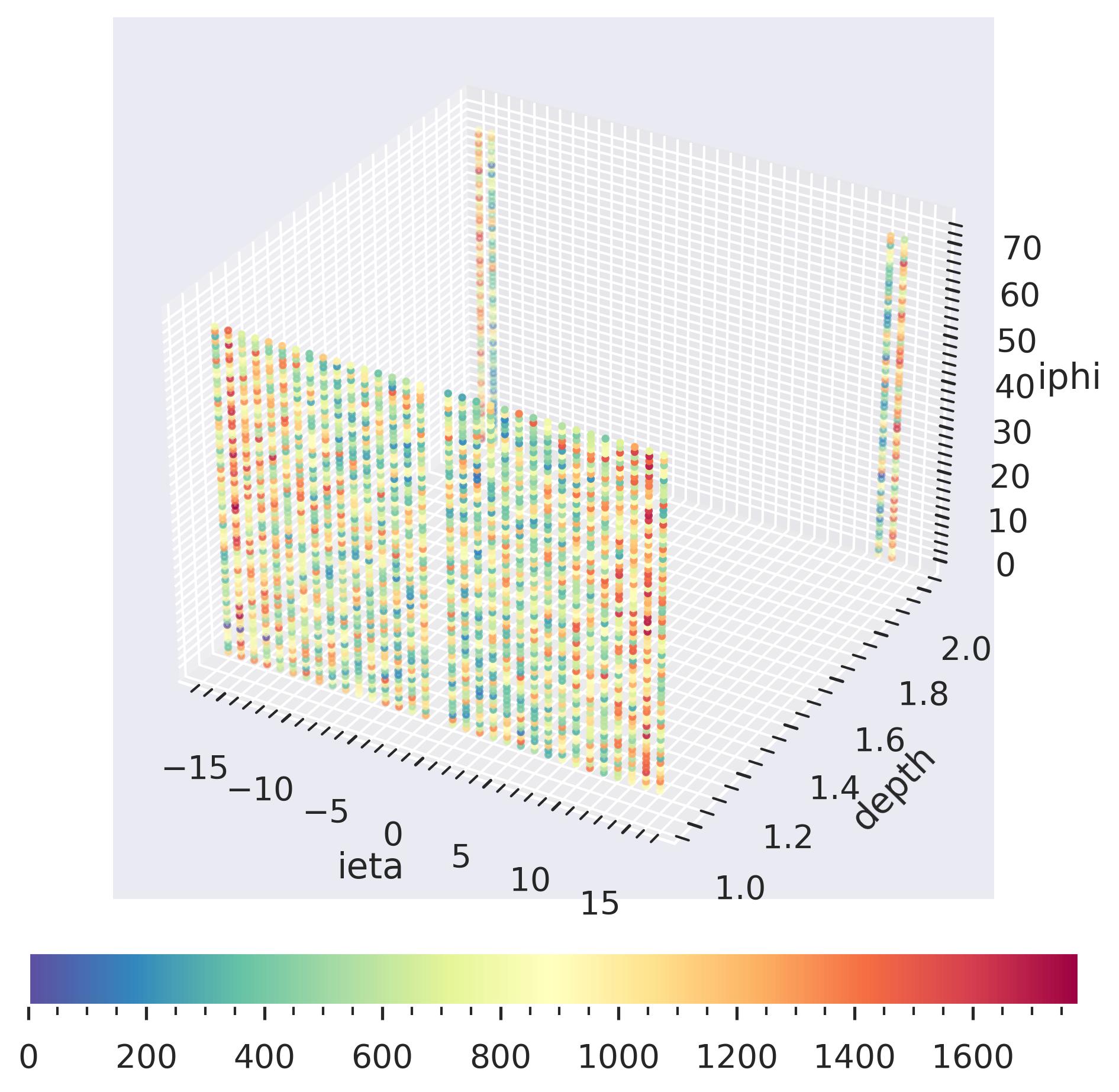}}
% \medskip
\end{adjustwidth}
\caption{A sample digi-occupancy map (\textit{year = 2018, RunId = 325,170, LS = 15)}: (\textbf{a}) digi-occupancy map for the HE and HB together; (\textbf{b}) the source system HE channels are placed in $\left| i\eta \right| \in [16, \dots, 29]$, $i\phi \in [1, \dots, 72]$, and~\emph{depth}~$\in [1, \dots, 7]$; and (\textbf{c}) the target system HB channels are placed in \mbox{$\left| i\eta \right| \in [1, \dots, 16]$}, $i\phi \in [1, \dots, 72]$, and~\emph{depth}~$\in [1, 2]$. The~HE and HB share similarities and differences in tasks, calorimeter technology, and~data characteristics. 
The missing sector at (\textbf{b}) corresponds to the two failed HE-RBX sectors during the 2018 collision runs.\label{fig:hehb_digioccupancy_sample}}
\end{figure}

We utilized the digi-occupancy data of the online DQM system of the CMS experiment to train and validate our models. 
The data contain healthy digi-occupancy maps with a 20~fC minimum threshold and were selected from certified good collision runs, as referred to by the ``Golden JSON'' of CMS. 
The digi-occupancy datasets were collected in 2018 during the LHC Run-2 collision experiment with a received luminosity per lumisection of up to $0.4$~pb$^{-1}$ and up to 2250 events. The~source and target datasets contain three-dimensional digi-occupancy maps for the HE and HB subsystems of the HCAL, respectively (as shown in Figure~\ref{fig:hehb_digioccupancy_sample}).

\vspace{-5pt}

The digi-occupancy map contains a particle hit count of the calorimeter readout channels for a given period of time. The~HCAL covers a considerable volume of CMS and has a fine segmentation along three axes ($i\eta \in [-32, \dots, 32]$, $i\phi \in [1, \dots, 72]$ and \emph{depth}~$\in [1, \dots, 7]$). The~$i\eta$ and $i\phi$ denote integer notation of the towers covering ranges of $\eta$ and $\phi$ of the CMS detector, respectively~\cite{collaboration2008cms}. The~digi-occupancy measurement corresponds to a hit record of the readout channels at the segmentation positions. 
The similarities between the source and target datasets and tasks have been established in the literature to be essential factors that impact the performance of TL~\cite{wen2019time}. 
The source system HE (as shown in Figure~\ref{fig:hehb_digioccupancy_sample}b) and the target system HB (as shown in Figure~\ref{fig:hehb_digioccupancy_sample}c) share a similar task but cover different segments of the HCAL.
Another major difference between the HE and HB in the 2018 LHC collision run is the front-end data acquisition optical-to-electrical technology, i.e.,~the HE was upgraded to SiPMs with QIE11 technology, and~the HB utilized HPD with QIE8. We compare the source and target datasets in Table~\ref{tbl:data_sets_desc}.

\begin{table}[H] \footnotesize
\centering
\caption{{Description} %MDPI: Please confirm the alignment change.
%Author: we accept the change.
 of source and target datasets.\label{tbl:data_sets_desc}}
\noindent
\begin{adjustwidth}{-\extralength}{0cm}
\begin{tabularx}{\fulllength}{cCCCcc}
\toprule
\textbf{Dataset} & \textbf{Sensor Technology} & \textbf{No. of Channels per RBX} & \textbf{No. of RBXes} & \textbf{Calorimeter Segmentation Ranges}                   & \textbf{Sample Size}                                                    \\ \midrule
Source (HE)                                 & SiPM                       & 192                              & 36                  & $\left| i\eta \right| \in [16, \dots, 29]$, $i\phi \in [1, \dots, 72]$, \emph{depth}~$\in [1, \dots, 7]$ & 20,000                      \\ %\hline
Target (HB)                                & HPD                        & 72                              & 36                  & $\left| i\eta \right| \in [1, \dots,16]$, $i\phi \in [1, \dots, 72]$, \emph{depth}~$\in [1, 2]$  & 7000                      \\ %\hline
\bottomrule
\end{tabularx}
\end{adjustwidth}
\end{table}
\unskip

\section{Methodology}
\label{sec:methodology}
This section presents the \textsc{GraphSTAD} modeling and the experimental setups for the transfer learning~study. 

\subsection{Data~Preprocessing}
This section describes the data preprocessing stages of the proposed approach, i.e.,~digi-occupancy renormalization and graph-adjacency matrix~generation.

\subsubsection{Digi-Occupancy Map~Renormalization}
\begin{figure}[H]
%\centering
%\captionsetup[subfigure]{justification=centering}
\includegraphics[width=1\textwidth, scale=1]{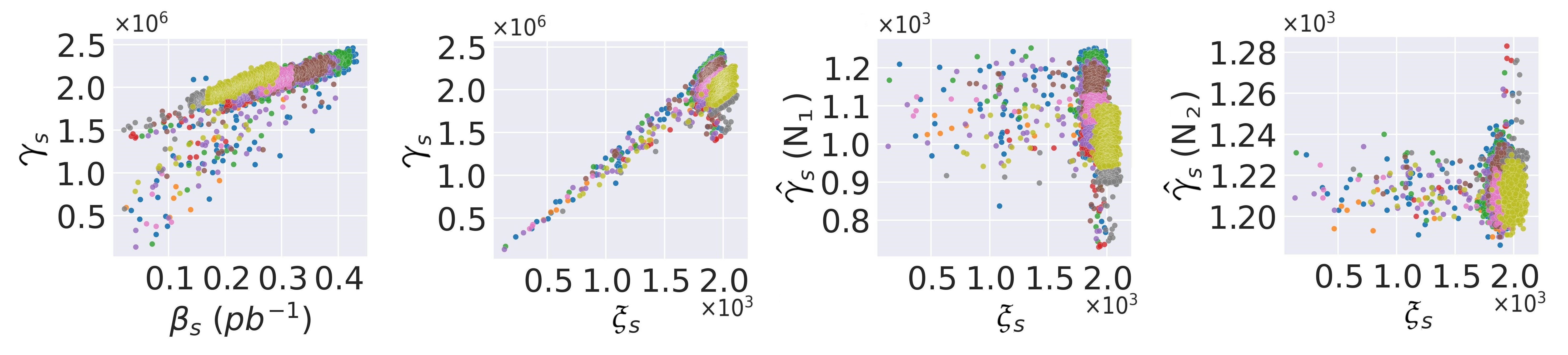}
\caption{{Total} %MDPI: Please add the explanations for the different colors in the figure.
% The explanation is given below in the last sentence of this figure caption.
 digi-occupancy data distribution of the HB and run settings per map ($s$): the received luminosity ($\beta_s$) and the number of events ($\xi_s$). $N_1$ is the renormalization of $\gamma_s$ based on $\xi_s$, and~$N_2$ is the reversible renormalization based on the median $\gamma$ along the $i\phi$ axis. 
The colors correspond to different collision runs.\label{fig:digioccuancy_RecLum_NumEvent_scatter}
}
\end{figure}

We apply digi-occupancy renormalization in the data preprocessing stages to normalize the values for the variation in the luminosity and the number of event configurations in the collision experiments~\cite{mulugeta2022dqm}.
The digi-occupancy ($\gamma$) map data of the HCAL vary with the received luminosity ($\beta$) and the number of events ($\xi$) (as shown in Figure~\ref{fig:digioccuancy_RecLum_NumEvent_scatter}). The~per-channel $\gamma _s (i)$ can range $\gamma _s (i) = [0, \xi_s]$, where $s$ denotes the $s^{\text{th}}$ LS (3D map) in the dataset and $i$ denotes the $i^{\text{th}}$ channel in the $s^{\text{th}}$ map. $\xi_s$ is usually adjusted with $\beta_s$, but not always. 
$\beta$ and $\gamma$ are retrieved from different systems on the existing CMS system; directly accessing $\beta$ for real-time $\gamma$ AD monitoring requires further effort. 
We renormalize the maps ($\gamma_s \rightarrow \hat{\gamma}_s$) based on $\xi_s$ to obtain consistent interpretation of the $\gamma$ maps across lumisections: 
\begin{linenomath}
\begin{equation}
\label{eq:stad_dqm__digioccp_norm}
    \hat{\gamma}_s = \frac{\gamma_s}{\xi_s}
\end{equation} 
\end{linenomath}

The renormalization of $\gamma$ with only $\xi$ does not entirely avoid the data distribution variations across collision runs, and~the distribution shifts and unpredictable spikes due to $\beta_s$ may affect the AD model training performance on the ST data (see Figure~\ref{fig:digioccuancy_RecLum_NumEvent_scatter}).
We employ additional reversible renormalization (RN) before and after invoking the AD model to mitigate the non-linearity of the $\gamma$ ST data. The~renormalization exploits the symmetric property of the $i\phi$ axis (the $\gamma$ channels are less diverse along the $i\phi$ axis); it divides the $\gamma$ channels per each $i\eta$ and $depth$ coordinate by the median values along the $i\phi$ axis on the model input and reverses the action on the model output. %EE: Please verify that intended meaning is retained
The remaining impact of $\beta_s$ is left to be learned by the AD model from the training~data. 

\unskip

\subsubsection{Adjacency Matrix~Generation}
We deployed an undirected graph network $\mathcal{G}(\mathcal{V}, \Theta)$ to represent the HCAL channels in a graph network based on their connections to a shared RBX system. The~graph $\mathcal{G}$ contains nodes $\upsilon \in \mathcal{V}$, with~edges $ (\upsilon _i, \upsilon _j) \in \Theta$ in a binary adjacency matrix $\mathcal{A} \in \mathbb{R}^{M \times M}$, where $M$ is the number of nodes (the channels). An~edge indicates the channels sharing the same RBX~as follows:
\begin{linenomath}
\begin{equation}
\label{eq:adjacency_mtrix}
    A(\upsilon _i, \upsilon _j) = 
\begin{cases}
   1,& \text{if } \Omega (\upsilon _i) = \Omega (\upsilon _j) \\
    0,              & \text{otherwise}
\end{cases}
\end{equation}
\end{linenomath}
where $\Omega (\upsilon)$ returns the RBX identification of the channel $\upsilon$.
There are approximately 7000 channels for the HE and 2600 for the HB in a graph representation of the digi-occupancy map. %EE: Please verify that intended meaning is retained
We retrieved the channel-to-RBX mapping from the 2018 HCAL's calorimeter segmentation~map.

\subsection{Anomaly Detection~Mechanism}

We denote the AE model of the AD system as $\mathcal{F}$. It takes ST data $\mathcal{X} \in \mathbb{R}^{T \times N_{i\eta} \times N_{i\phi} \times N_d \times N_f}$ as a sequence in a time window $ t_x \in [t-T, t]$, where $N_{i\eta} \times N_{i\phi} \times N_d$ is the spatial dimension corresponding to the $i\eta$, $i\phi$, and~$depth$ axes, respectively, and~$N_f$ is the number of input variables ($N_f=1$, as we monitored only a digi-occupancy quantity in the spatial data). The~$\mathcal{F}_{\theta, \omega}: \mathcal{X} \to \bar{\mathcal{X}}$, parametrized by $\theta$ and $\omega$, attempts to reconstruct the input ST data $\mathcal{X}$ and outputs $\bar{\mathcal{X}}$.  
The encoder network of the model $\mathcal{E}_\theta: \mathcal{X} \to \mathcal{Z}$ provides low-dimension latent space, $\mathcal{Z} = \mathcal{E}_\theta(\mathcal{X})$, and~the decoder $\mathcal{D}_\omega: \mathcal{Z} \to \bar{\mathcal{X}}$ reconstructs the ST data from $\mathcal{Z}$, $\bar{\mathcal{X}}=\mathcal{D}_\omega(\mathcal{Z})$ as follows: 
\begin{linenomath}
\begin{equation}
\bar{\mathcal{X}} = \mathcal{F}_{\theta, \omega}(\mathcal{X}) = \mathcal{D}_\omega(\mathcal{E}_\theta(\mathcal{X}))
\end{equation}
\end{linenomath}

Anomalies can live for a short time on a single digi-occupancy map, or~they can persist over time, affecting a sequence of maps. Aggregated spatial reconstruction error is calculated over a time window $T$ using mean absolute error (MAE) to capture a time-persistent anomaly as follows:
\begin{linenomath}
\begin{equation}
e_{i, MAE} = \frac{1}{T} \sum^t_{t^\prime=t-T}|x_i (t^\prime)-\bar{x}_i (t^\prime)|
\end{equation}
\end{linenomath}
where $x_i \in \mathcal{X}$ and $\bar{x}_i \in \bar{\mathcal{X}}$ are the input and the reconstructed digi-occupancy of the {$i\text{th}$} %MDPI: We removed the italics and superscript of “th”, please confirm.
%Author: we accept the change
 channel.
We standardized $e_{i, MAE}$ to homogenize the reconstruction accuracy variations among the channels when generating the anomaly score $a_i$ as follows:
\begin{linenomath}
\begin{equation}
a_i = \frac{e_{i, MAE}}{\sigma _i}
\end{equation}
\end{linenomath}
where $\sigma _i$ is the standard deviation of the $e_{i, MAE}$ on the training dataset. 
The standardized anomaly score allows us to use a single AD decision threshold $\alpha$ for all the channels in the spatial map.
The anomaly flags are generated after applying $\alpha$ to the anomaly scores ($a_i > \alpha$).
The $\alpha$ value can be tuned to control the detection~sensitivity.

\begin{figure}[H]
\centering
\includegraphics[width=1\textwidth, scale=1]{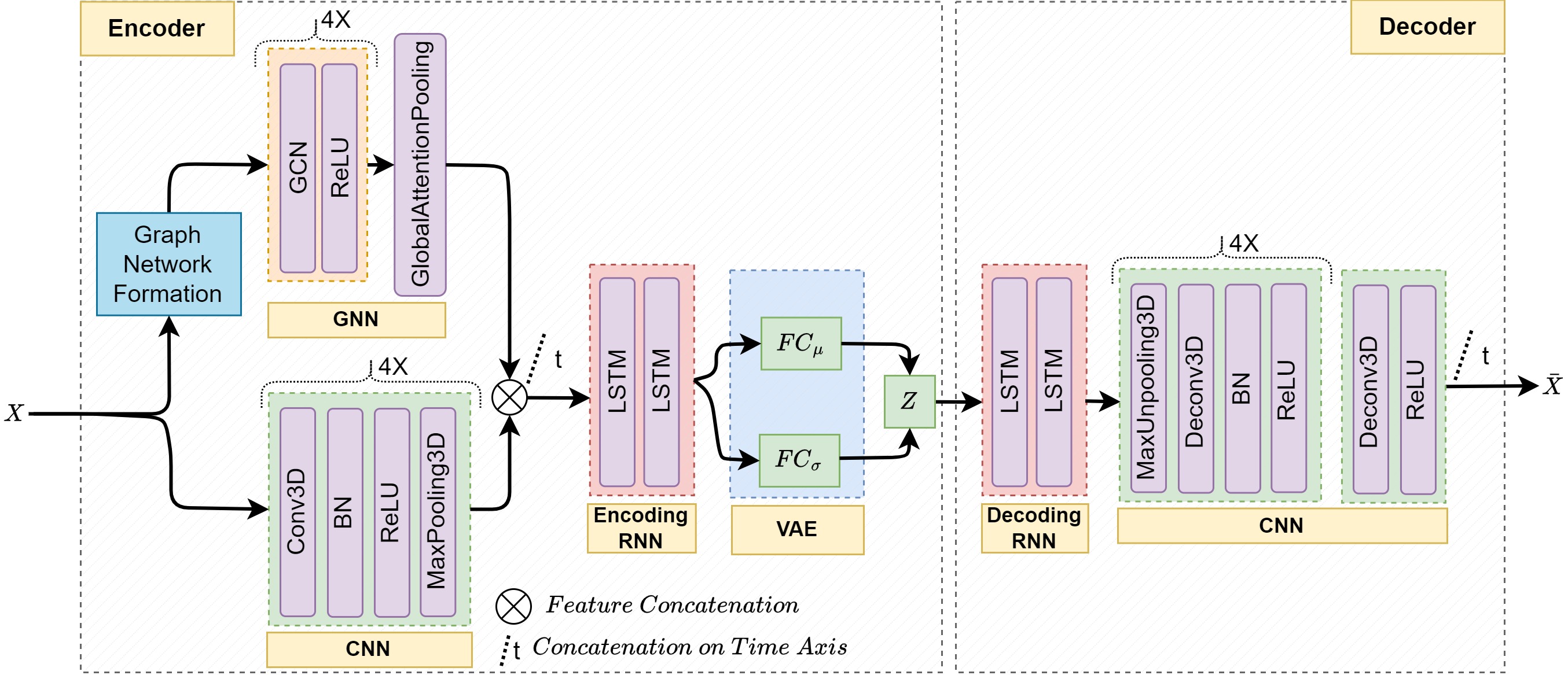}\footnotesize{\par
Conv3D: 3D convolutional neural network; GCN: graph convolutional neural networks; Deconv3D: 3D deconvolutional neural networks; BN: batch normalization; LSTM: long short-term memory recurrent networks; FC: fully connected neural networks; VAE: variational AE.
}\par
\caption{The architecture of the proposed AE for the \textsc{GraphSTAD} system~\cite{mulugeta2022dqm}.~The~GNN and CNN provide spatial feature extraction for each time step, %EE: Please verify that intended meaning is retained
and~the RNN network captures the temporal behavior of the extracted features. The~feature extraction $\mathcal{E}_\theta$ incorporates the GNN for back-end physical connectivity among the spatial channels, CNN for regional spatial proximity of the channels, and~RNN for temporal behavior extraction. 
$\mathcal{D}_\omega$ contains RNNs and deconvolutional neural networks to reconstruct the ST input data from the low-dimensional latent features. \label{fig:propose_autoencoder_detailed_diagram}
}
\end{figure}

The use-case \textsc{GraphSTAD} AE model is made of CNN, GNN, and~RNN networks;
it employs a CNN and GNN with a pooling mechanism to extract relevant features from spatial DQM data followed by RNN to capture the temporal characteristics of the extracted features (see Figure~\ref{fig:propose_autoencoder_detailed_diagram}).
It integrates a variational layer~\cite{kingma2013auto} at the end of the encoder for regularization of overfitting by enforcing continuous and normally distributed latent representations~\cite{asres2021unsupervised, mulugeta2022dqm}. 
We refer readers to ref.~\cite{mulugeta2022dqm} for further discussion of the mathematical formulation and architecture of the \textsc{GraphSTAD} model.

We trained the AE on healthy digi-occupancy maps (without significant anomaly contamination, see Section~\ref{sec:datasetdescription}) of the target HB system. 
We normalized the spatial data per channel into a $[0, 1]$ range to train the model across the variations in calorimeter channels effectively. 
We utilized a mean squared error (MSE) loss function as follows:
\begin{linenomath}
\begin{equation}
\label{eq:weighted_training_loss}
\mathcal{L}_{MSE} = \frac{1}{M}\sum_{i}(x_i -\bar{x}_i)^2
\end{equation}
\end{linenomath}
where $x_i$ and  $\bar{x}_i$ are the input and the reconstructed values of the normalized $\hat{\gamma}$ of the $i^{\text{th}}$ channel, respectively, and~$M$ is the total number of channels.
The variational layer of the AE (denoted as VAE in Figure~\ref{fig:propose_autoencoder_detailed_diagram}) regularizes the training MSE loss using the \textit{Kullback--Leibler divergence} (KL) distance $D_{KL}$~\cite{kingma2013auto} to achieve the normally distributed latent space as follows: 
\begin{linenomath}
\begin{equation}
\label{eq:training_loss}
\mathcal{L} = \underset{W \in \mathbb{R}} {\text{argmin}}
\left\{ \mathcal{L}_{MSE} - \lambda {D_{KL}}\left[ \mathcal{N}(\mu_z, \sigma_z),  \mathcal{N}(0, I)\right] + \rho \|W\|_{2}^{2} \right\}
\end{equation}
\end{linenomath}
where $\mathcal{N}$ is a normal distribution with zero mean and unit variance, and~$\|.\|_{2}^{2}$ is a squared  \textit{Frobenius norm} of $L_2$ \textit{regularization} for the trainable model parameters $W$~\cite{van2017l2}. \mbox{$\lambda=0.003$ and $\rho=10^{-7}$} are tunable regularization hyperparameters.
We employed the \textit{Adam} optimizer~\cite{kingma2014adam} for~training.

\subsection{Transfer Learning~Approach}

Model parameter TL generally consists of four basic steps: (1) selection of a source task with a related modeling problem and an abundance of data where we can exploit the mapping knowledge from the inputs to outputs, (2) development of the source model that performs well in the source task, (3) transfer source model to target model where whole or part of the source model is employed as part of the target model, and (4) fine-tuning the target model on the target dataset if necessary.
We present knowledge transfer on \textsc{GraphSTAD} AE models, i.e.,~an AD model trained on digi-occupancy maps of the source HE subsystem is transferred to the target HB subsystem. 
Direct transfer of knowledge from the source into the target, irrespective of their divergence and thorough investigation of the model network layers, would limit the efficacy of TL in the target domain~\cite{shao2014transfer, yu2022survey, chato2023survey}. 

\begin{figure}[H]
\centering
\includegraphics[width=\textwidth, scale=1]{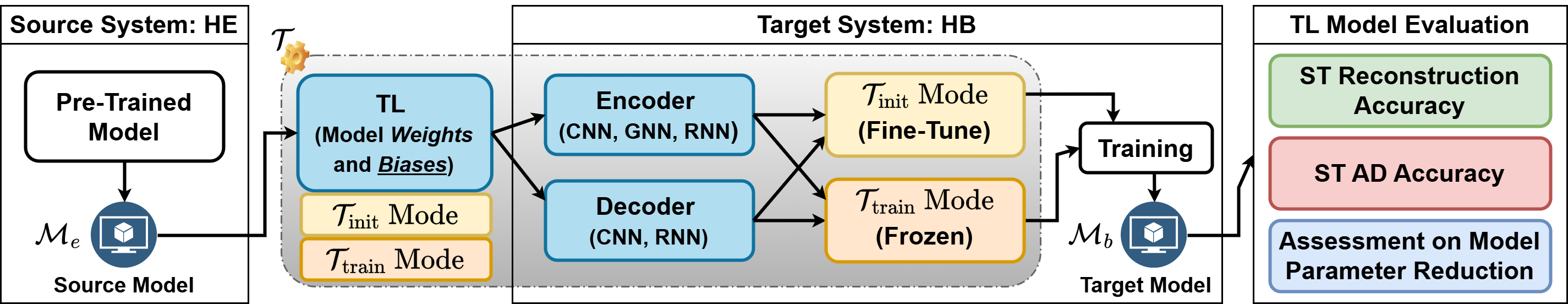}
\caption{{{General} %MDPI: Please confirm if the dashed arrows and solid arrows need explanations in the figure. If yes, please add.
% There is not much significance difference b/n the dashed arrow and solid arrows. We used dashes just to highlight that multiple configurations are supported by the lines.  We have updated the figure, making all solid lines to avoid confusion.
 framework of the proposed transfer learning mechanism. }}
\label{fig:tl_diagram}
\end{figure}

{We have thus investigated several transferring cases when employing TL in two principal model training phases: initialization and training (see Figure~\ref{fig:tl_diagram})}.
\begin{itemize}
    \item {Init mode} ($\mathcal{T}_{\text{init}}$): the trainable network parameters (weights and bias) of the source model are transferred into the target model initialization. 
    The target model is further trained on the target HB dataset, resulting in fine-tuning.
    \item {Train mode} ($\mathcal{T}_{\text{train}}$): The model parameters of the source model are directly reused as the final inference parameters of the target model; the parameters are frozen and excluded from fine-tuning on the target HB dataset.
\end{itemize}

Let $\mathcal{M}(\Psi,\Omega)$ be an AD model with parameters $\Psi$ and $\Omega$ that represent the model networks that can be affected and not affected by TL, respectively. $\mathcal{M}_e(\Psi_e,\Omega_e)$ and $\mathcal{M}_b(\Psi_b,\Omega_b)$ are the source and target models for the HE and HB, respectively. The~TL modes of $\mathcal{T}$ can be formulated mathematically as follows:
\begin{linenomath}
\begin{equation}
\begin{split}
     \mathcal{T}_{\text{init}} &: \mathcal{M}_b (\Psi_e,\Omega_b) \underset{\text{fine-tuning}}{\rightarrow } \mathcal{M}_b (\Psi_e^{'}, \Omega_b^{'} )\\
     \mathcal{T}_{\text{train}} &: \mathcal{M}_b (\Psi_e,\Omega_b) \underset{\text{fine-tuning}}{\rightarrow } \mathcal{M}_b (\Psi_e, \Omega_b^{'} ) \\
    \end{split}
\end{equation}
\end{linenomath}
where the superscript $^{'}$ denotes the parameters that are updated after fine-tuning the $\mathcal{M}_b$ model on the target~dataset.

$\mathcal{F}_{\theta, \omega}$ of the \textsc{GraphSTAD} is made of CNNs and GNNs with a pooling mechanism to extract relevant features from high-dimensional spatial data, followed by RNNs to capture the temporal characteristics of the extracted features (as shown in Figure~\ref{fig:propose_autoencoder_detailed_diagram}).
Table~\ref{tbl:tl_configs} presents the TL mechanisms that we apply to the different deep networks of the encoder and decoder to study the impacts on ST digi-occupancy map reconstruction and AD accuracy. 
We also analyze the effects of RNN state preservation within and across time windows.
We further investigate variations in training iterations and learning rate scheduling methods. 
The discussion includes the impact of the TL on reconstruction and AD accuracy, saturation, training stability, and~the reduction in the number of model trainable~parameters. 

The implementation of parameter transferring on DL networks can be accomplished in two ways: (1) start with the source model and then reset (remove and add) the networks that are not included in the TL, and~(2) start with the target model with random initialization and update the parameter values of the networks included in the TL from the corresponding source networks. 
The first approach is widely utilized in DL TL literature and employed for feature extraction; however, it may not be suitable for flexibly choosing layers at different hierarchies, as~the target models might have slight variations. 
Several configuration setups of the AE are derived from the spatial configuration of the input 3D map, which differs for the source HE and target HB systems---e.g., variation in the depth spatial dimension between HE and HB. 
We have found the second approach more convenient for our study, as~we intend to apply TL on different networks of the encoder and decoder of the AE~model. 

\newlength{\kw}
\setlength{\kw}{6.4cm}
\begin{table}[H] \small
\caption{{Transfer} %MDPI: Please confirm the alignment change.
%Authors: we confirm and accept the change
 learning experiment~configurations.\label{tbl:tl_configs}}
\noindent
\begin{adjustwidth}{-\extralength}{0cm}
\resizebox{1.34\textwidth}{!}{
\begin{tabularx}{\fulllength}{ccCcC}
\toprule
\multirow{2.5}{*}{\textbf{Config.}} & \multicolumn{2}{c}{\textbf{Init Mode ($\mathbf{\mathcal{T}_{init}}$)}} & \multicolumn{2}{c}{\textbf{Train Mode ($\mathbf{\mathcal{T}_{train}}$)}}                                                                                      \\  \cmidrule{2-5} 
                                     & \multicolumn{1}{c}{\textbf{Notation}}                                                           & \multicolumn{1}{c}{\textbf{Description}}                                                                                                                                                       & \multicolumn{1}{c}{\textbf{Notation}}                                     & \multicolumn{1}{c}{\textbf{Description}}                                                  \\ \midrule
1                                    & \multicolumn{1}{c}{RANDOM}                                                                      & $\mathcal{M}_b$ is initialized randomly  (weights: using \textsc{Kaiming} uniform~\cite{he2015delving}, and~biases: zero)                                                                                                                                    & \multicolumn{1}{c}{No-TL}                                                 & 
 Complete training (fine-tuning)                           \\  \midrule
2                                    & \multicolumn{1}{c}{\multirow{5}{*}{\centering TL-4}}          & \multirow{5}{*}{\parbox{\kw}{\centering $\mathcal{M}_d$ is initialized randomly, except~the spatial learning networks (CNN and GNN) are initialized by TL from $\mathcal{M}_e$}}                                 & \multicolumn{1}{c}{No-TL}                                                 &     Complete training (fine-tuning)                                                                    \\ 
3                                    & \multicolumn{1}{c}{}                                                                            &                                                                                                                                                                            & \multicolumn{1}{c}{TL-1}                               & GNN of $\mathcal{E}_\theta$ is frozen (not fine-tuned)                     \\ 
4                                    & \multicolumn{1}{c}{}                                                                            &                                                                                                                                                                            & \multicolumn{1}{c}{TL-2}                               & CNN of $\mathcal{E}_\theta$ is~frozen                                      \\ 
5                                    & \multicolumn{1}{c}{}                                                                            &                                                                                                                                                                            & \multicolumn{1}{c}{TL-2$_d$}                               & CNN of $\mathcal{D}_\omega$ is~frozen                                      \\ 
6                                    & \multicolumn{1}{c}{}                                                                            &                                                                                                                                                                            & \multicolumn{1}{c}{TL-3}                          & CNN and GNN of $\mathcal{E}_\theta$ are~frozen                             \\ 
\midrule
7                                   & \multicolumn{1}{c}{\multirow{3}{*}{\centering TL-7}} & \multirow{2}{*}{\parbox{\kw}{\centering All the spatial and temporal learning networks (CNN, GNN, and~RNN) of the $\mathcal{M}_b$ are initialized by TL from $\mathcal{M}_e$}} & \multicolumn{1}{c}{TL-5}                     & CNN, GNN, and RNN of $\mathcal{E}_\theta$ are~frozen                        \\ 
8                                    & \multicolumn{1}{c}{}                                                                            &                                                                                                                                                                            & \multicolumn{1}{c}{TL-6} & CNN, GNN, and RNN of $\mathcal{E}_\theta$, and~RNN of $\mathcal{D}_\omega$ are frozen     \\ %\hline
\bottomrule
\end{tabularx}
}
\end{adjustwidth}
\noindent{\footnotesize{{{TL}%MDPI: Please check all the bold in the whole table footer and confirm if they are unnecessary and can be removed.
% The bold was added to enhance legibility, but we removed it as it is unnecessary.
}: transfer learning is applied.  
{TL-1}: ENCODER{[}GNN{]}, 
{TL-2}: ENCODER{[}CNN{]}, 
{TL-2$_d$}: DECODER{[}CNN{]}, 
{TL-3}: ENCODER{[}CNN, GNN{]}, 
{TL-4}: ENCODER{[}CNN, GNN{]}, DECODER{[}CNN{]}, 
{TL-5}: ENCODER{[}CNN, GNN, RNN{]}, 
{TL-6}: ENCODER{[}CNN, GNN, RNN{]}, DECODER{[}RNN{]}, 
{TL-7}: ENCODER{[}CNN, GNN, RNN{]}, DECODER{[}CNN, RNN{]}.
}
}
\end{table}
\unskip

\section{Results and~Discussion}
\label{sec:resultsanddiscussion}

This section will discuss the results of TL on different network layers of the \textsc{GraphSTAD} AE models. 
We will investigate the effects of TL on reconstruction accuracy, trainable parameter reduction, and~AD performance on the target HB digi-occupancy map dataset. 
We applied TL for model initialization ($\mathcal{T}_{\text{init}}$) and~training or fine-tuning ($\mathcal{T}_{\text{train}}$) on the target HB dataset. 
We trained the models on NVIDIA Tesla V100 with 4 GPUs using 4000~digi-occupancy maps from LS 1 to 500 and evaluated them on a test set that contains approximately 3000 maps from LS from 500 to 1500. 
We utilized 20\% of the training dataset for validation loss calculation during training to determine the best states for the models. 
We set the LR at $10^{-3}$ to train the models with five LSs per time~window. 

\subsection{Spatio-Temporal Reconstruction~Performance}

We will discuss below the reconstruction performance (using $\mathcal{L}_{MSE}$) of TL applied on spatial (CNNs and GNNs) and temporal (RNNs) learning networks. 
We will also briefly present comparison results for the LR scheduling~choice.

\subsubsection{Transfer Learning on Spatial Learning~Networks}
\label{sec:tl_spatial}

\begin{figure}[H]

%\captionsetup[subfigure]{justification=centering}
\begin{adjustwidth}{-\extralength}{0cm}\centering
\subfloat[\centering]{\includegraphics[width=0.86\textwidth]{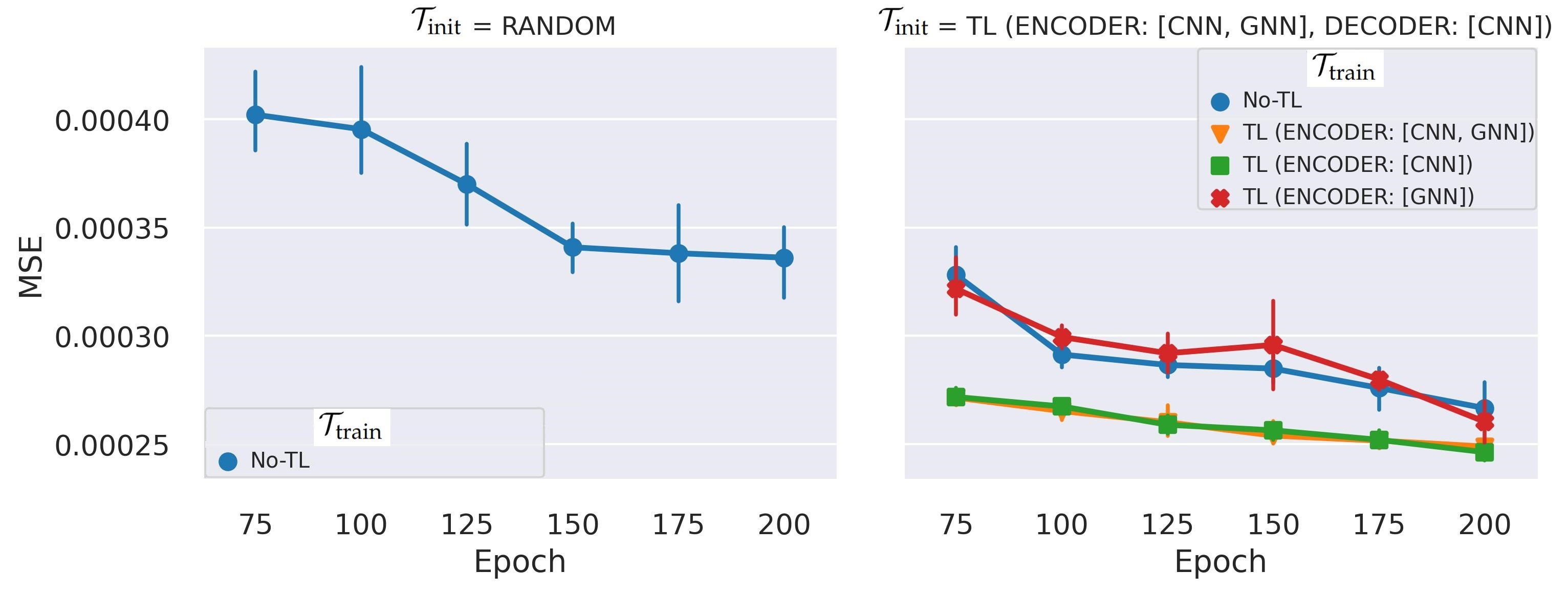}}
% \medskip
\subfloat[\centering]
{\includegraphics[width=0.47\textwidth]{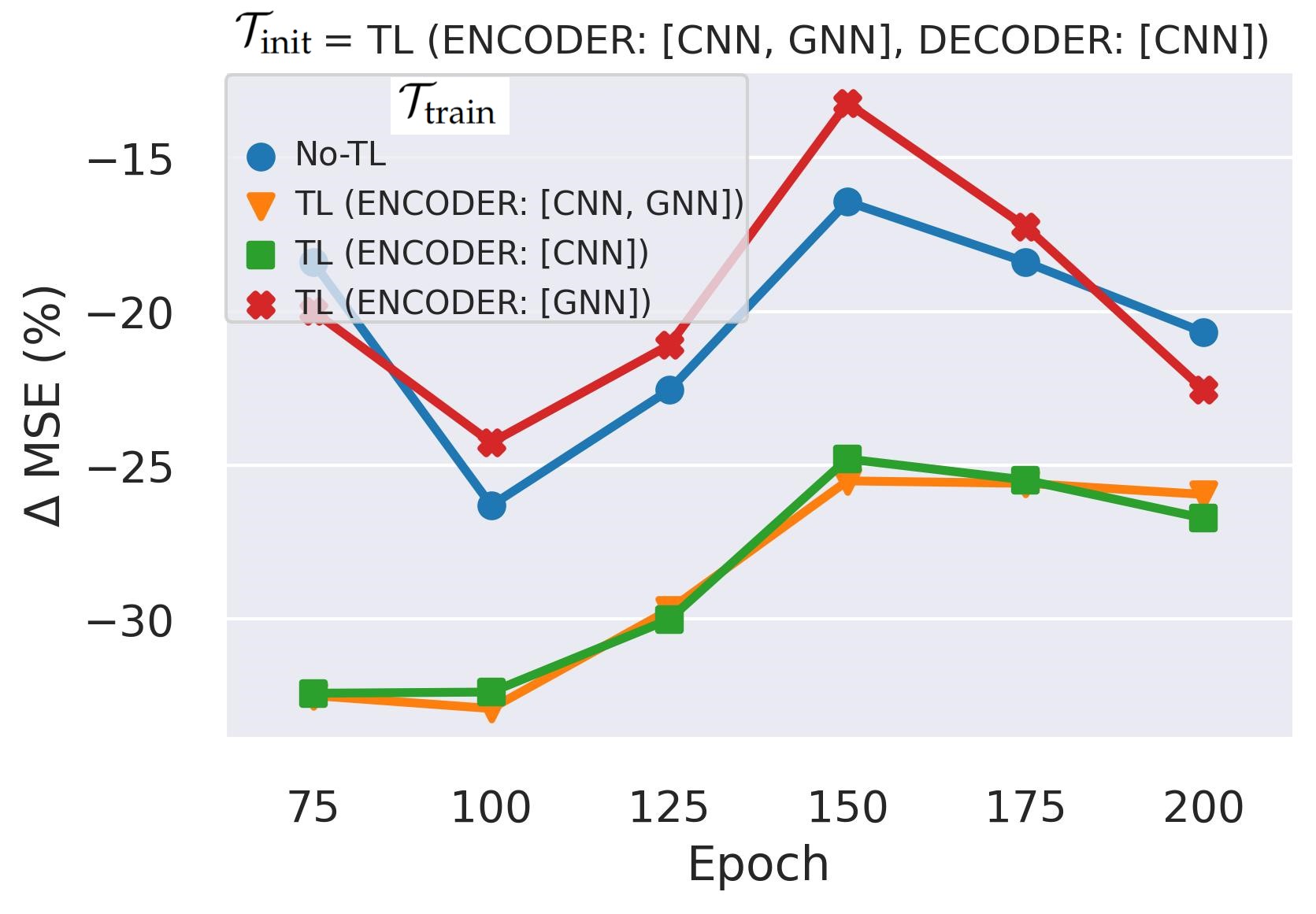}}
% \medskip
\end{adjustwidth}
\caption{{Reconstruction} %MDPI: There is no yellow, green and red dots in the left image of subfigure (a) but exist explanations, please check and revise the extra legends.
%Authors: we have removed the extra legends and updated the figure accordingly
 $\mathcal{L}_{MSE}$ performance of the TL on spatial networks across different epochs: (\textbf{a}) test MSE loss, where the bars show the dispersion of five repeated experiments; (\textbf{b}) the average relative difference between the MSE loss and no-TL. TL is applied with $\mathcal{T}_{\text{init}}$ on the encoder and decoder using TL-4: ENCODER{[}CNN, GNN{]}, DECODER{[}CNN{]}, and~$\mathcal{T}_{\text{train}}$ on the encoder using TL-1: ENCODER{[}GNN{]}, TL-2: ENCODER{[}CNN{]}, and~TL-3: ENCODER{[}CNN, GNN{]}. The~no-TL model starts to saturate at $epoch>150$. \label{fig:ZeroBias__2018__nocut_HB__epochs_vs_test_loss_multi_epoch_compare}}
\end{figure}

We have assessed the transferability of the DL AD model at the initialization and inference phases for the spatial learning networks (CNNs and GNNs) on both the encoder and decoder networks on different numbers of training epochs (see Figure~\ref{fig:ZeroBias__2018__nocut_HB__epochs_vs_test_loss_multi_epoch_compare}). 
The TL has reduced the reconstruction error $\mathcal{L}_{MSE}$ of healthy maps by 32.5\% to 20.7\% when the number of epochs is varied from 75 to 200 (as shown in Figure~\ref{fig:ZeroBias__2018__nocut_HB__epochs_vs_test_loss_multi_epoch_compare}b). 
The minimum gain of 13\% is achieved at epoch 150, just before the performance of the no-TL model starts to saturate.
The complete fine-tuning---TL for initialization, followed by fine-tuning the whole network---provided around 20\% improvement. 
The $\mathcal{L}_{MSE}$ generally decreases, while the relative TL gain roughly decreases as the epoch increases to 150. 
The results are not entirely unexpected; the DL models may improve performance as the training epoch increases, reducing the gap caused by the difference in the initialization and training mechanisms. 
When the epoch increased beyond 150, the~randomly initialized model using no-TL achieves only slight improvement, whereas the $\mathcal{L}_{MSE}$ continues to drop for the TL models, increasing the relative gain of the TL. 
The initialization TL on all the spatial learning networks of the AE using $\mathcal{T}_{\text{init}}$~=~TL-4 and training only the decoder while freezing the encoder using $\mathcal{T}_{\text{train}}$~=~TL-3 achieves the best improvement, from~26\% to 32.5\%. 
The TL gain of the GNNs is limited compared to CNNs; the CNNs are the primary networks that learn the input spatial data, and they have 15 times more parameters than the GNNs in the use-case \textsc{GraphSTAD} AE model. 
Transferring and freezing the CNNs of the encoder (TL-2 and TL-3) results in stable performance on repeated experiments. Although TL-2 outperforms TL-3 slightly (by $3\%$) at epoch 200, TL-3 provides computational leverage ($7.2\%$, see Section~\ref{sec:tl_st}), bypassing the training overhead of the~GNNs.

\vspace{-5pt}

Table~\ref{tbl:recon_loss_compare_epoch_200} further provides the average and best ST reconstruction performance at $epoch = 200$. 
Inference TL on the decoder networks without fine-tuning \mbox{$\mathcal{T}_{\text{train}}$~=~TL-2$_d$} fails to reconstruct the target data adequately.
In an AE architecture, the~encoder maps the input into low-dimensional latent space (information compression), while the decoder attempts to reconstruct (information expansion) the target data from the latent space. The~decoder networks thus require fine-tuning on the target dataset to adjust their parameters to the target reconstruction effectively. 
Boulle~et~al.~\cite{boulle2020classification} investigated TL and DL for a univariate chaotic time series classification model; they argued that BN without fine-tuning limits the transferability of CNNs.
The scaling and shifting parameters for BN and bias parameters are estimated from the training dataset and strongly correlate to the data. 
We further studied TL on the decoder when the BN layer and the bias parameters of the CNNs are fine-tuned on the target dataset. 
$\mathcal{L}_{MSE}$ is substantially improved by 50\% compared to the frozen decoder (see Table~\ref{tbl:recon_loss_decoder_bn_bias_compare_epoch_200}). 
However, the~error is still 20 times higher than that without TL, indicating that the CNNs of the decoder also require fine-tuning to achieve reasonable accuracy. 
The results demonstrate the promising leverage of TL for AE model initialization on both feature extraction encoder and reconstruction decoder networks, whereas fine-tuning with the target dataset is essential for the decoder~networks. 

\begin{table}[H] %\footnotesize
\caption{ST reconstruction $\mathcal{L}_{MSE}$ of TL on spatial networks ($epoch=200$). \label{tbl:recon_loss_compare_epoch_200}}
%\resizebox{0.9\textwidth}{!}{
\begin{tabularx}{\textwidth}{CCCC}
\toprule
\multicolumn{1}{c}{$\mathbf{\mathbold{\mathcal{T}_{\text{init}}}}$} & \multicolumn{1}{c}{$\mathbf{\mathcal{T}_{train}}$} & \multicolumn{1}{c}{$\mathbf{\mathcal{L}_{MSE}\downarrow}$}                & \textbf{$\mathbf{\Delta\mathcal{L}_{MSE}}$ w.r.t $\mathbf{\mathcal{T}_{init}}$=$\mathbf{RANDOM}$$\downarrow$} \\\midrule
\multicolumn{4}{c}{{{Average Score} %MDPI: Please confirm if the bold formatting is necessary; if not, please remove it. The following highlights are the same.
%Author: we have removed the bold
}}                                                                      \\\midrule
RANDOM                                             & No-TL               & $3.361 \times 10^{-4}$          & --                                                                          \\ 
TL-4                                               & No-TL               & $2.666 \times 10^{-4}$          & $-$20.7\%                                                                     \\ 
TL-4                                               & TL-1                & $2.604 \times 10^{-4}$          & $-$22.5\%                                                                     \\ 
TL-4                                               & TL-2                & $\mathbf{2.463 \times 10^{-4}}$ & \textbf{$-$26.7\%}                                                            \\ 
TL-4                                               & TL-3                & $2.489 \times 10^{-4}$          & $-$25.9\%                                                                     \\ 
TL-4                                               & TL-2$_d$            & $1.530 \times 10^{-2}$          &  4452.2\%                          \\ \midrule
\multicolumn{4}{c}{{{Best Score}}}                                                                             \\\midrule
RANDOM                                             & No-TL               & $3.085 \times 10^{-4}$          & --                                                                          \\ 
TL-4                                               & No-TL               & $2.569 \times 10^{-4}$          & $-$16.7\%                                                                     \\ 
TL-4                                               & TL-1                & $2.502 \times 10^{-4}$          & $-$18.9\%                                                                     \\ 
TL-4                                               & TL-2                & $\mathbf{2.420 \times 10^{-4}}$ & \textbf{$-$21.6\%}                                                            \\ 
TL-4                                               & TL-3                & $2.451 \times 10^{-4}$          & $-$20.5\%                                                                     \\ 
TL-4                                               & TL-2$_d$            & $1.5255 \times 10^{-2}$         &  4844.9\%                          \\ 
\bottomrule
\end{tabularx}
%}
\noindent{\footnotesize{{TL-1:} %MDPI: We merged the table footer into one paragraph, please confirm.
%Author: we confirm the change
 ENCODER{[}GNN{]}, 
TL-2: ENCODER{[}CNN{]}, 
TL-2$_d$: DECODER: {[}CNN{]}, 
TL-3: ENCODER{[}CNN, GNN{]}, 
TL-4: ENCODER{[}CNN, GNN{]}, DECODER{[}CNN{]}. 
The \textbf{{bold font}} is the best score and, the down arrow ($\downarrow$) indicates that lower is better.}}
\end{table}
\unskip

\begin{table}[H] \small
\caption{ST reconstruction $\mathcal{L}_{MSE}$ of TL for spatial networks ($\mathcal{T}_{\text{init}}$=$\operatorname{TL-4}$, $epoch=200$, average score). \label{tbl:recon_loss_decoder_bn_bias_compare_epoch_200}}
%\resizebox{0.9\textwidth}{!}{
\begin{tabularx}{\textwidth}{CCCC}
\toprule
$\mathbf{\mathcal{T}_{init}}$ & $\mathbf{\mathcal{T}_{train}}$ & $\mathbf{\mathcal{L}_{MSE}\downarrow}$               & \textbf{$\mathbf{\Delta\mathcal{L}_{MSE}}$ w.r.t TL-$2_d$$\downarrow$} \\\midrule
TL-4 & TL-2$_d$                                  & $1.530 \times 10^{-2}$          & --                                  \\ 
TL-4 & TL-2$_d$/[BN]              & ${7.200 \times 10^{-3}}$ & {$-$53.0\%}                    \\ 
TL-4 & TL-2$_d$/[BN, BIAS]         & $7.354 \times 10^{-3}$          & $-$51.9\%                             \\ 
\bottomrule
\end{tabularx}
%}
\noindent{\footnotesize{TL-2$_d$: DECODER{[}CNN{]}, 
TL-4: ENCODER{[}CNN, GNN{]}, DECODER{[}CNN{]}, 
and / denotes excluding.
}
}
\end{table}
\unskip

\subsubsection{Transfer Learning on Spatio-Temporal Learning~Networks}
\label{sec:tl_st}

We investigated TL on the temporal RNNs (LSTM layers) in both the encoder and decoder networks, along with the spatial learning networks (CNNs and GNNs), using $\mathcal{T}_{\text{init}}$~=~TL-4 and $\mathcal{T}_{\text{train}}$~=~TL-3---the~best performing TL for spatial networks across epochs (see Figure~\ref{fig:ZeroBias__2018__nocut_HB__epochs_vs_test_loss_multi_epoch_compare}). %EE: Please verify that intended meaning is retained

Table~\ref{tbl:recon_loss_parmas_compare_best_models_all} presents $\mathcal{L}_{MSE}$ when TL is applied to the ST networks. We evaluated the models by preserving the RNN states across time windows that leverage the accuracy. 
When the TL involves freezing the RNNs of the decoder ($\mathcal{T}_{\text{train}}$~=~TL-6), the~$\mathcal{L}_{MSE}$ improves by $22.6$--$32.6\%$ while considerably reducing the model trainable parameters by $97.77\%$, mainly due to the frozen LSTM networks (see Table~\ref{tbl:recon_loss_parmas_compare_best_models_all}).

\begin{figure}[H]

%\captionsetup[subfigure]{justification=centering}
\begin{adjustwidth}{-\extralength}{0cm}\centering
\subfloat[\centering]{\includegraphics[width=0.65\textwidth]{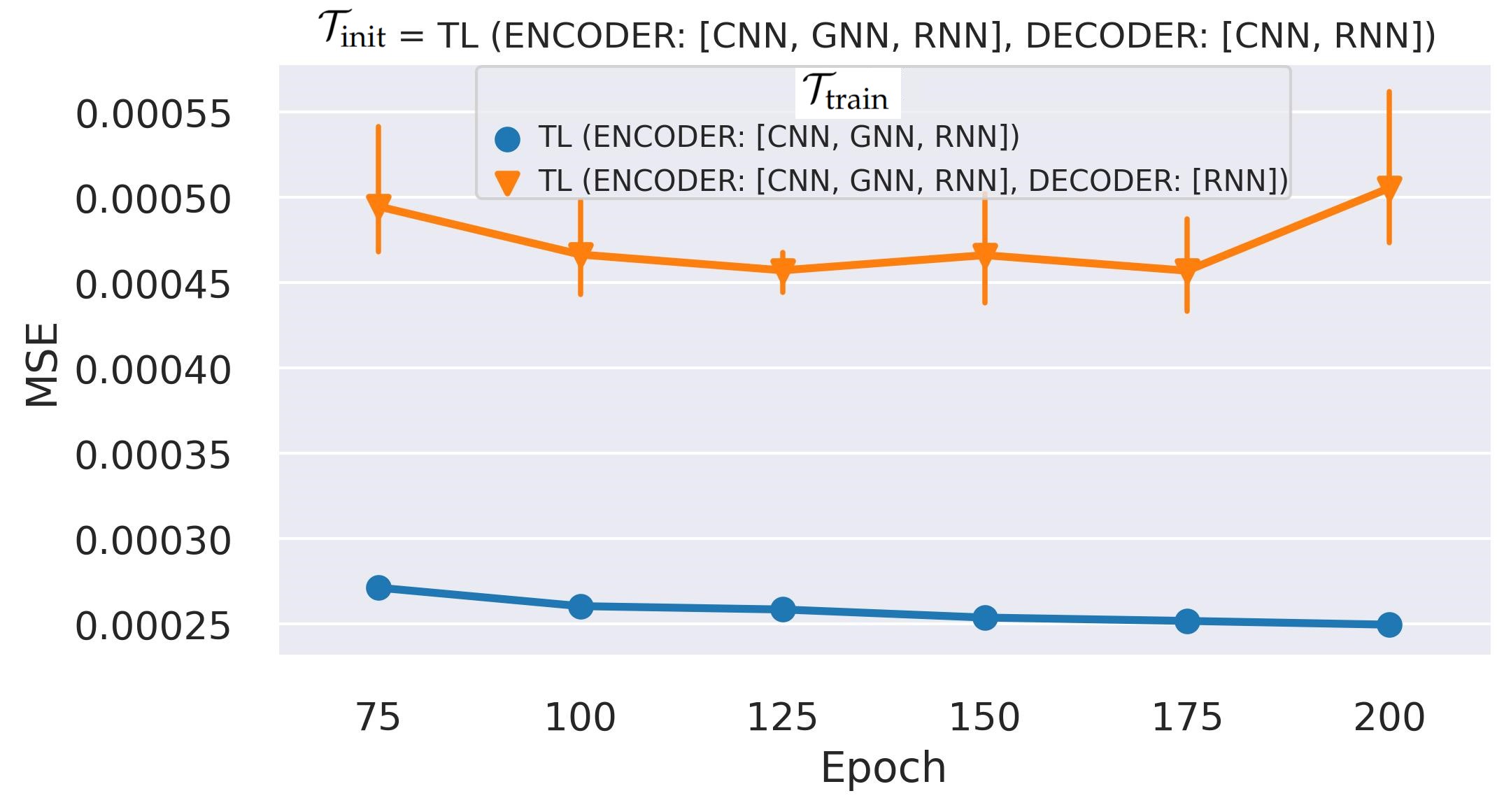}}
% \medskip
\subfloat[\centering]{\includegraphics[width=0.61\textwidth]{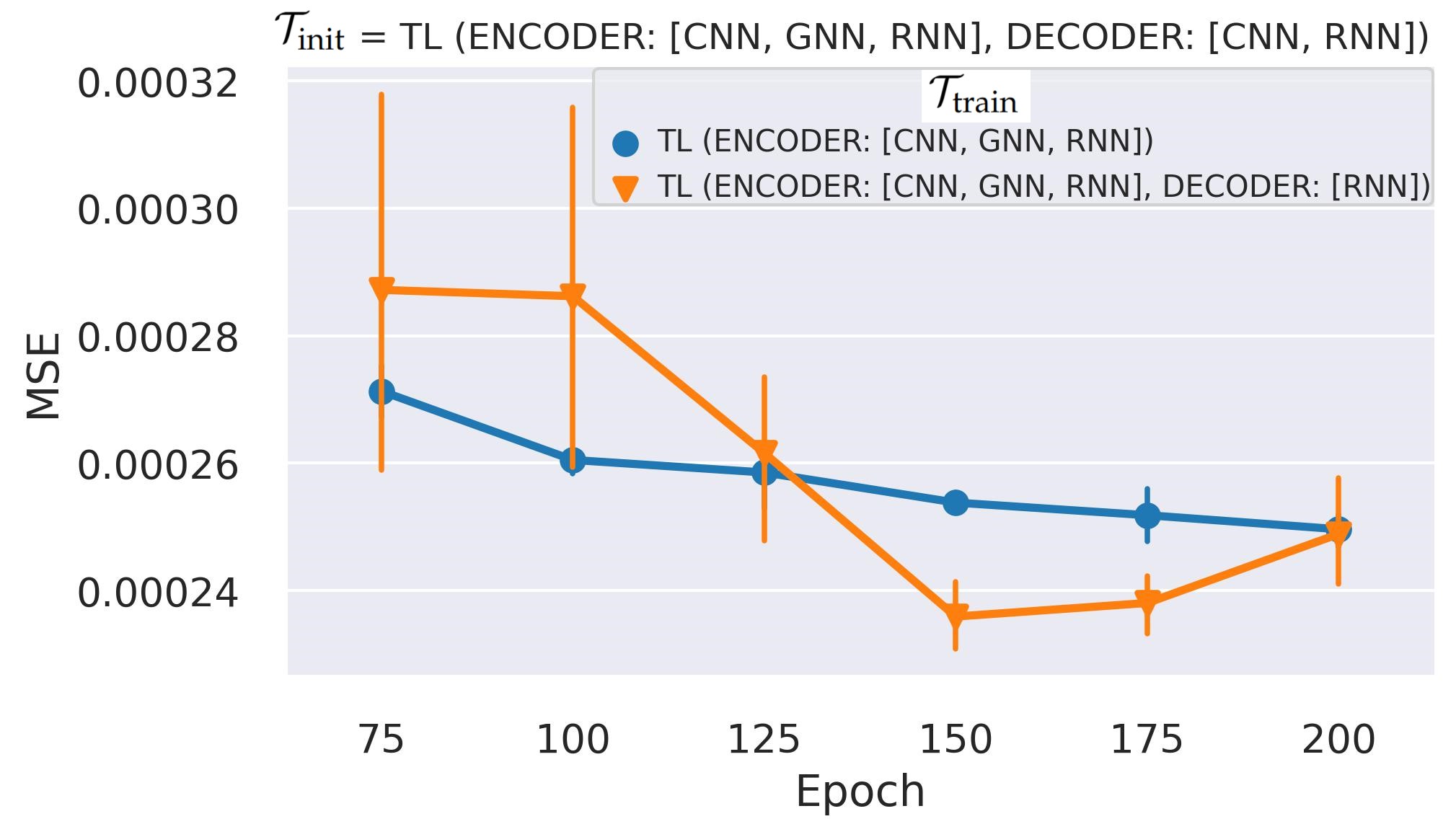}}
% \medskip
\end{adjustwidth}
\caption{Reconstruction $\mathcal{L}_{MSE}$ performance of TL on the ST networks. 
TL is applied to train the encoder and decoder with TL-5: (ENCODER: {[}CNN, GNN, RNN{]}) and~TL-6: (ENCODER: {[}CNN, GNN, RNN{]}, DECODER: {[}RNN{]}). 
The MSE loss in (\textbf{a}) non-preserved LSTM states that reset for each time window, and~(\textbf{b}) preserved LSTM states across consecutive time windows.
The bars show the dispersion of five repeated experiments. Lower epochs have higher performance variations among repeated experiments, and~the stabilization is better at higher epochs.}
\label{fig:ZeroBias__2018__nocut_HB__epochs_vs_test_loss_multi_epoch_compare_rnn_}
\end{figure}

\begin{figure}[H]

%\captionsetup[subfigure]{justification=centering}
\begin{adjustwidth}{-\extralength}{0cm}\centering
\subfloat[\centering]{\includegraphics[width=0.55\textwidth]{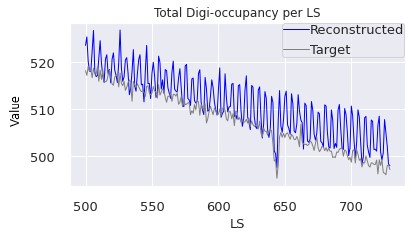}}
% \medskip
\subfloat[\centering]
{\includegraphics[width=0.55\textwidth]{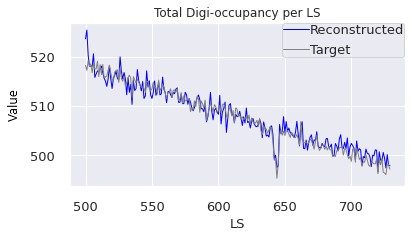}}
% \medskip
\end{adjustwidth}
\caption{\textls[-15]{Digi-occupancy map reconstruction on sample ST data from the test set. The~model was trained using TL-6, and~the inference was executed (\textbf{a}) without and (\textbf{b}) with LSTM state preservation across time windows. The~AE operates on ST $\gamma$ maps, but~the curves in these plots correspond to the aggregate renormalized $\gamma$ per LS to illustrate the model's performance in handling the fluctuation across~lumisections}.\label{fig:total_digioccpuancy_per_ls_TL_RNN_vs_state_preserve_bn_TWs}}
\end{figure}

However, the~performance of $\mathcal{T}_{\text{train}}$~=~TL-6 suffers substantially, increasing $\mathcal{L}_{MSE}$ by more than 50\% if~the state memory of the RNNs is not preserved across the sliding time window, i.e.,~memory reset at every non-overlapping sliding time window start (as shown in Figure~\ref{fig:ZeroBias__2018__nocut_HB__epochs_vs_test_loss_multi_epoch_compare_rnn_}). 
Figure~\ref{fig:ZeroBias__2018__nocut_HB__epochs_vs_test_loss_multi_epoch_compare_rnn_} presents the $\mathcal{L}_{MSE}$ values on multiple epochs when the TL includes the RNNs with and without state preservation across time windows. 
The plots show a significant enhancement by preserving the states on the frozen decoder RNNs using $\mathcal{T}_{\text{train}}$~=~TL-6 but~a limited impact when the target dataset fine-tunes the decoder RNNs using \mbox{$\mathcal{T}_{\text{train}}$~=~TL-5.}  
Figure~\ref{fig:total_digioccpuancy_per_ls_TL_RNN_vs_state_preserve_bn_TWs}a demonstrates that the TL-6 model struggles to reconstruct the map at the first time step in each sliding time window when states are not preserved across time windows. 
This is caused by the model's reliance solely on the input map for the first time-step reconstruction with reset memory states (zeros), while the states are adjusted and improved for the subsequent maps. 
The reconstruction improves when utilizing previous states, even for the first maps in the time windows (see Figure~\ref{fig:total_digioccpuancy_per_ls_TL_RNN_vs_state_preserve_bn_TWs}b). 

\begin{table}[H] \scriptsize
\caption{{ST} %MDPI: We moved Table 5 after its first citation. Please confirm.
%Author: we confirm the change
 reconstruction $\mathcal{L}_{MSE}$ of TL on ST~networks.\label{tbl:recon_loss_parmas_compare_best_models_all}}
%\resizebox{0.9\textwidth}{!}{
\begin{tabularx}{\textwidth}{ccCCC}
\toprule
$\mathbf{\mathcal{T}_{init}}$                            & $\mathbf{\mathcal{T}_{train}}$                    & $\mathbf{\mathcal{L}_{MSE}\downarrow}$  & \textbf{$\mathbf{\Delta\mathcal{L}_{MSE}}$ w.r.t $\mathbf{\mathcal{T}_{init}}$=$\mathbf{RANDOM}\downarrow$} & \textbf{$\mathbf{\Delta\mathcal{W}}$ w.r.t $\mathbf{\mathcal{T}_{init}}$=$\mathbf{RANDOM}\downarrow$} \\ \midrule
\multicolumn{5}{c}{{{Best Score at ${epoch=75}$}}} \\ \midrule

 RANDOM                                                     & No-TL                                                 & $3.826 \times 10^{-4}$                        & --                                &  \\ 
 TL-4           & No-TL                                                 & $3.180 \times 10^{-4}$                        & $-$16.9\%                               & 0.00\%  \\ 
 TL-4           & TL-1                               & $3.082 \times 10^{-4}$                        & $-$19.5\%                               & $-$0.17\% \\ 
 TL-4           & TL-2                               & $2.686 \times 10^{-4}$                        & $-$29.8\%                               & $-$2.23\% \\ 
 TL-4           & TL-3                          & $2.705 \times 10^{-4}$                        & -29.3\%                               &  $-$2.39\% \\ 
 TL-7 & TL-5                     & $2.667 \times 10^{-4}$                        & $-$30.3\%                               & $-$8.38\%  \\ 
 TL-7 & TL-6 & {$\mathbf{2.577 \times 10^{-4}}$} %MDPI: Please add an explanation for the use of bold on the numbers in the whole table in the table footer. If the bold is unnecessary, please remove them. 
%Author: We have added the explanation in the footnote.
                        & \textbf{$-$32.6\%}                               & \textbf{$-$97.77\%}    \\ \midrule

\multicolumn{5}{c}{{{Best Score at ${epoch=200}$}}} \\ \midrule
 RANDOM                                                     & No-TL                                                 & $3.085 \times 10^{-4}$                             & --                                     & --                                                          \\ 
 TL-4           & No-TL                                                 & $2.569 \times 10^{-4}$                             & $-$16.7\%                                    & 0.00\%                                                          \\ 
 TL-4           & TL-1                               & $2.502 \times 10^{-4}$                             & $-$18.9\%                                    & $-$0.17\%                                                         \\ 
 TL-4           & TL-2                               & $2.420 \times 10^{-4}$                             & $-$21.6\%                                    &$-$2.23\%                                                         \\ 
 TL-4           & TL-3                          & $2.451 \times 10^{-4}$                             & $-$20.5\%                                    & $-$2.39\%                                                         \\ 
 TL-7 & TL-5                     & $2.457 \times 10^{-4}$                             & $-$20.4\%                                    & $-$8.38\%                                                         \\ 
 TL-7 & TL-6  & $\mathbf{2.389 \times 10^{-4}}$	                       & \textbf{$-$22.6\% }                                 & \textbf{$-$97.77\%}                                              \\ 
 \bottomrule
\end{tabularx}
%}
\noindent{\footnotesize{{TL-1:} %MDPI: We merged the table footer into one paragraph. Please confirm.
%%Author: We confirm the change.
 ENCODER{[}GNN{]}, 
TL-2: ENCODER{[}CNN{]}, 
TL-3: ENCODER{[}CNN, GNN{]}, 
TL-4: ENCODER{[}CNN, GNN{]}, DECODER{[}CNN{]}, 
TL-5: ENCODER{[}CNN, GNN, RNN{]}, 
TL-6: ENCODER{[}CNN, GNN, RNN{]}, DECODER{[}RNN{]}, 
TL-7: ENCODER{[}CNN, GNN, RNN{]}, DECODER{[}CNN, RNN{]}.
{$\mathbf{\Delta\mathcal{W}}$} is the reduction in the number of trainable model~parameters. 
The \textbf{{bold font}} is the best score and, the down arrow ($\downarrow$) indicates that lower is better.
}}
\end{table}
\unskip

\vspace{-9pt}
\unskip

\vspace{-5pt}
%\unskip
%
%
%\unskip

\subsubsection{Applying Learning Rate~Scheduling}

The AE $\mathcal{L}_{MSE}$ reaches saturation after $epoch > 150$ when trained without TL (as illustrated in Figure~\ref{fig:ZeroBias__2018__nocut_HB__epochs_vs_test_loss_multi_epoch_compare}a). 
Learning rate (LR) scheduling mechanisms, e.g.,~lowering the LR when the loss flattens, or fast convergence methods, could mitigate training stagnation. 
We have investigated the impact of scheduling on the TL by training the model with super-convergence \textit{one-cyclic} LR scheduling~\cite{smith2019super}. 
The LR scheduling sets the LR according to a one-cycle policy that anneals the LR from an initial LR ($init\_lr=4\times10^{-5}$) to a maximum LR ($max\_lr=10^{-3}$) and then from that maximum LR to a minimum LR ($min\_lr=4\times10^{-7}$). 
We utilize a cosine annealing mechanism along with the other settings of the scheduler, such as $div\_factor=25$ and~$final\_div\_factor=100$, where $div\_factor$ determines the initial LR by dividing $max\_lr$, and $final\_div\_factor$ estimates $min\_lr$ by dividing the initial LR.
We have kept the default values of the remaining hyperparameters given in the {{PyTorch=1.12.0} %MDPI: 1. Please state which version of the software was used. 2. Please confirm if the small caps font is necessary, if not, please revise.
%Author: 1) we have added the PyTorch version and 2) we have removed the small caps
} implementation~\cite{smith2019super}.

Table~\ref{tbl:recon_loss_parmas_compare_lr_sch} shows that the LR scheduling has improved the $\mathcal{L}_{MSE}$ compared to the fixed LR (provided in Table~\ref{tbl:recon_loss_parmas_compare_best_models_all}) by 19\% for without TL, and~7.1\% and 4.3
\% for TL with \mbox{$\mathcal{T}_{\text{train}}$~=~TL-5}, and~$\mathcal{T}_{\text{train}}$~=~TL-6, respectively. 
The relative progress of the TL is approximately 9\% with the LR scheduling, which is lower than the 22.6\% achieved with the fixed LR. 
The results are consistent with Figure~\ref{fig:ZeroBias__2018__nocut_HB__epochs_vs_test_loss_multi_epoch_compare}b, showing a narrowing of the performance difference as the number of epochs increases past $epoch > 150$, with performance saturating for the model without TL, $\mathcal{T}_{\text{init}}$~=~$\operatorname{RANDOM}$. %EE: Please verify that intended meaning is retained
The cyclic LR scheduling method may require more configuration tuning effort to improve the performance compared to fixed LR or other simpler LR scheduling~approaches.

\begin{table}[H] \small
\tablesize{}
\caption{ST reconstruction $\mathcal{L}_{MSE}$ of TL with LR scheduling mechanism ($epoch=200$, best score).\label{tbl:recon_loss_parmas_compare_lr_sch}}
%\resizebox{0.9\textwidth}{!}{
\begin{tabularx}{\textwidth}{CCCC}
\toprule
$\mathbf{\mathcal{T}_{init}}$                            & $\mathbf{\mathcal{T}_{train}}$                    & $\mathbf{\mathcal{L}_{MSE}\downarrow}$  & \textbf{$\mathbf{\Delta\mathcal{L}_{MSE}}$ w.r.t $\mathbf{\mathcal{T}_{init}}$=$\mathbf{RANDOM}\downarrow$}  \\ \midrule

RANDOM                                                     & No-TL                                                 & $2.500 \times 10^{-4}$                        & --                               \\ 
TL-4           & No-TL                                                 & $2.400 \times 10^{-4}$                        & $-$4.0\%                                \\ 
TL-4 & TL-3                     & $2.460 \times 10^{-4}$                        & $-$1.6\%                                \\ 
TL-7 & TL-5                     & {$\mathbf{2.283 \times 10^{-4}}$} %MDPI: Please add an explanation for the use of bold on the numbers in the whole talbe in the table footer. If the bold is unnecessary, please remove them. 
%Authors: we added the explanation at the footnote
                        & \textbf{$-$8.7\%}                                \\ 
TL-7 & TL-6 & ${2.286 \times 10^{-4}}$               & {$-$8.6\%}                                                   \\ 
\bottomrule
\end{tabularx}
%}

\noindent{\footnotesize{
TL-3: ENCODER{[}CNN, GNN{]}, 
TL-4: ENCODER{[}CNN, GNN{]}, DECODER{[}CNN{]}, 
TL-5: ENCODER{[}CNN, GNN, RNN{]}, 
TL-6: ENCODER{[}CNN, GNN, RNN{]}, DECODER{[}RNN{]}, 
TL-7: ENCODER{[}CNN, GNN, RNN{]}, DECODER{[}CNN, RNN{]}.
The \textbf{{bold font}} is the best score and, the down arrow ($\downarrow$) indicates that lower is better.
}}
\end{table}
\unskip

\subsection{Anomaly Detection~Performance}
\label{sec:perfwithcounters}

Machine learning studies performed thus far in the CMS DQM system have primarily employed simulated anomaly data to evaluate the efficacy of the developed AD \mbox{models~\cite{azzolin2019improving, mulugeta2022dqm}}; a small fraction of the DQM data is affected by real anomalies that are inadequate for comprehensive model validation.
We validate the AD models on synthetic anomalies simulating real channel anomalies of the HCAL~\cite{mulugeta2022dqm}. We have generated synthetic anomalies simulating \textit{dead}, \textit{hot}, and~\textit{degraded} channels and injected them into healthy digi-occupancy maps of the test dataset. We formulate the simulated channel anomalies as follows:
\begin{linenomath}
\begin{equation}
    \gamma_{a} = 
\begin{cases}
    R_D \gamma _{h}, & \text{where}~R_D \neq 1~\text{and}~\gamma_{a} \leq \xi\\
    \gamma_{a} = \xi, & \text{where}~R_D = 1~\text{and}~\gamma _{h} < \xi
\end{cases}
\end{equation}
\end{linenomath}
where $\gamma_{a}\in [0,\xi]$ and $\gamma_{h}\in (0,\xi]$ are the digi-occupancy of the generated anomaly channel and its corresponding expected healthy reading, respectively. The~$R_D$ is the degradation factor, and~the channel anomalies are defined as follows:\vspace{6pt}
\begin{linenomath}
\begin{equation}
\label{eq:anml_type_gen}
\begin{split}  
 \text{Dead} & : \gamma_{a} =0,~\text{using}~R_D=0 \\
 \text{Degraded} & : 0 < \gamma_{a}  < \gamma _{h},~\text{using}~0<R_D<1 \\
 \text{Noisy hot} & :  \gamma _{h} < \gamma_{a} \leq \xi,~\text{using}~R_D>1\\
 \text{Fully hot} & : \gamma _{h} < \gamma_{a} = \xi,~\text{using}~R_D=1\\
\end{split}
\end{equation}
\end{linenomath}

The algorithm that generates the synthetic anomaly samples involves three steps~\cite{mulugeta2022dqm}: (1) selection of a random set of LSs from the test set, (2) random selection of spatial locations $\varphi$ for each LS, where $\varphi\in [i\eta \times i\phi \times$~\emph{depth}$]$ on the HB axes (see Figure~\ref{fig:hehb_digioccupancy_sample}c), and~(3) injection of the simulated anomalies into digi-occupancy maps of the LSs.
The simulated anomalies include {dead}, {degraded}, {noisy hot}, and~{fully hot} channels. 
For consistency, we have kept the same spatial locations for all the anomaly types. 
We have evaluated the performance on several classification metrics using three anomaly thresholds set to capture 90\%, 95\%, and~99\% of the injected~anomalies.

We evaluate the AD accuracy on 14,000 digi-occupancy maps (2000 maps for each anomaly type) for the dead ($R_D = 0\%$), decaying anomalies ($R_D = [80\%, 60\%, 40\%, 20\%]$), noisy hot ($R_D=200\%$), and~fully-hot ($\gamma_ a = \xi$) channels. We investigate persistent channel anomalies that affect consecutive maps in a time window. We have processed 70,000 digi-occupancy maps (with the generated 1.17\% abnormal channels) that include five history maps in the time window for each of the 14,000~maps.

We compare the AD performance of models without TL and the best TL from Table~\ref{tbl:recon_loss_parmas_compare_lr_sch}. The~models are denoted as follows: 
\begin{linenomath}
\begin{equation}
\begin{split}  
    \mathcal{M}_{\text{N}} & : \mathcal{T}_{\text{init}}~=~\operatorname{RANDOM}~\text{and}~\mathcal{T}_{\text{train}}~=~\operatorname{No-TL} \\ \mathcal{M}_{\text{T}} & : \mathcal{T}_{\text{init}}~=~\operatorname{TL-7}~\text{and}~\mathcal{T}_{\text{train}}~=~\operatorname{TL-6} \\
\end{split}
\end{equation}
\end{linenomath}
where $\mathcal{M}_{\text{N}}$ and $\mathcal{M}_{\text{T}}$ are the models trained without TL and with TL, respectively.

\begin{figure}[H]

\captionsetup[subfigure]{justification=centering}
\begin{adjustwidth}{-\extralength}{0cm}\centering
\subfloat[\centering]{\includegraphics[width=1.34\textwidth]{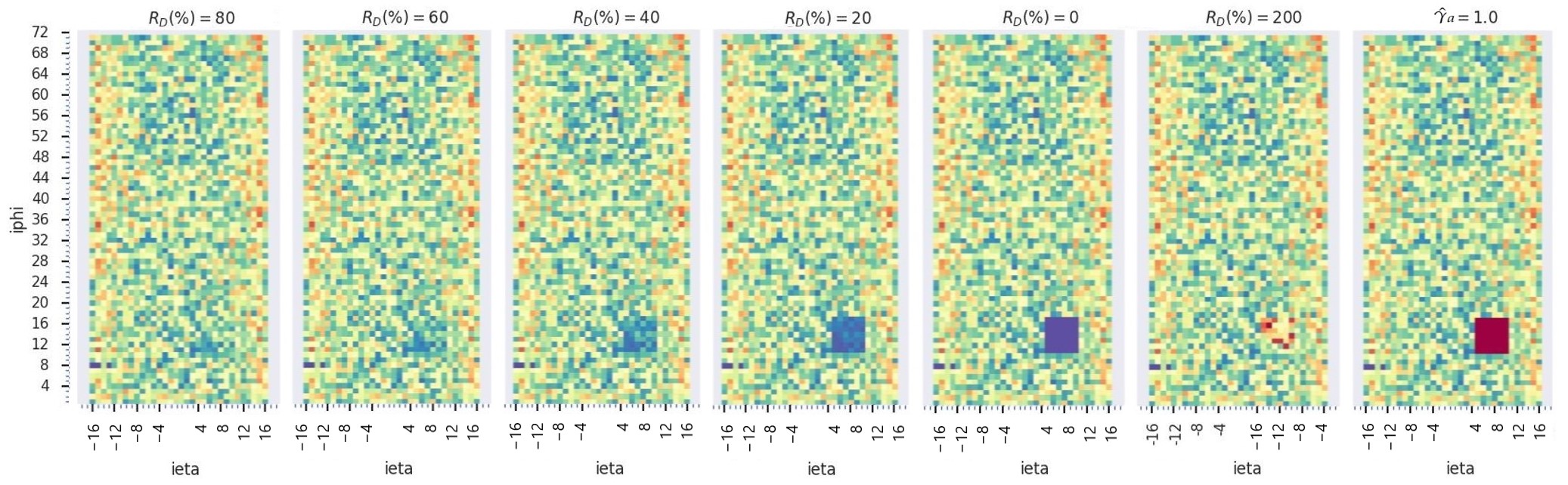}}
% \medskip
 \\
\subfloat[\centering]{\includegraphics[width=1.34\textwidth]{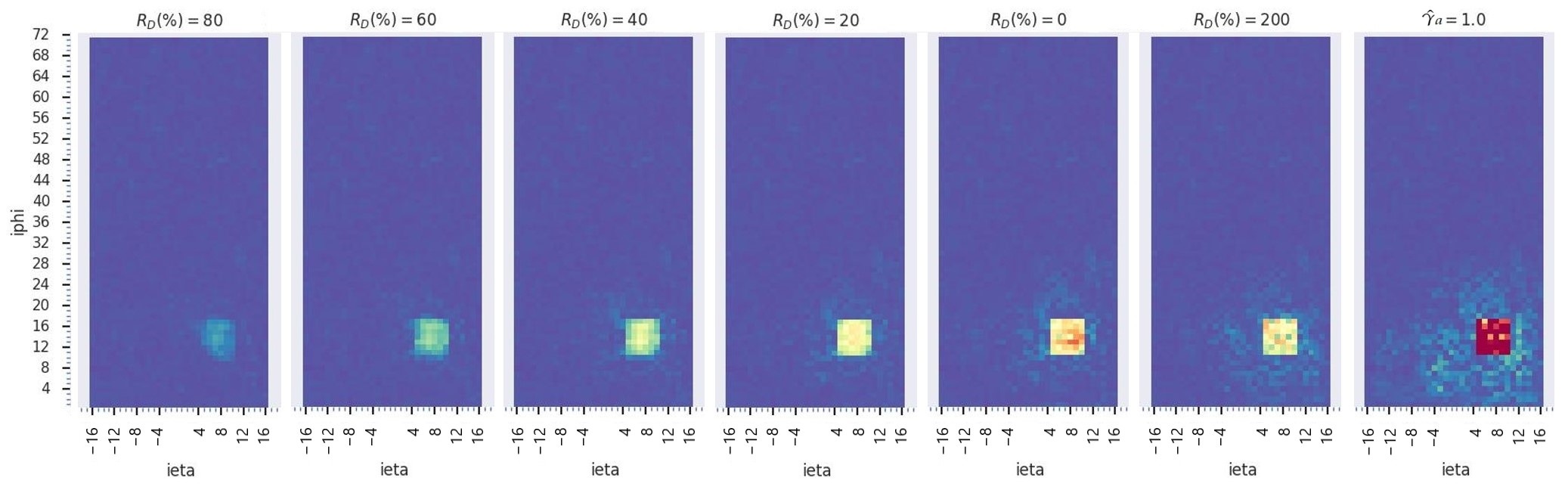}}
% \medskip
 \\
\subfloat[\centering]{\includegraphics[width=1.34\textwidth]{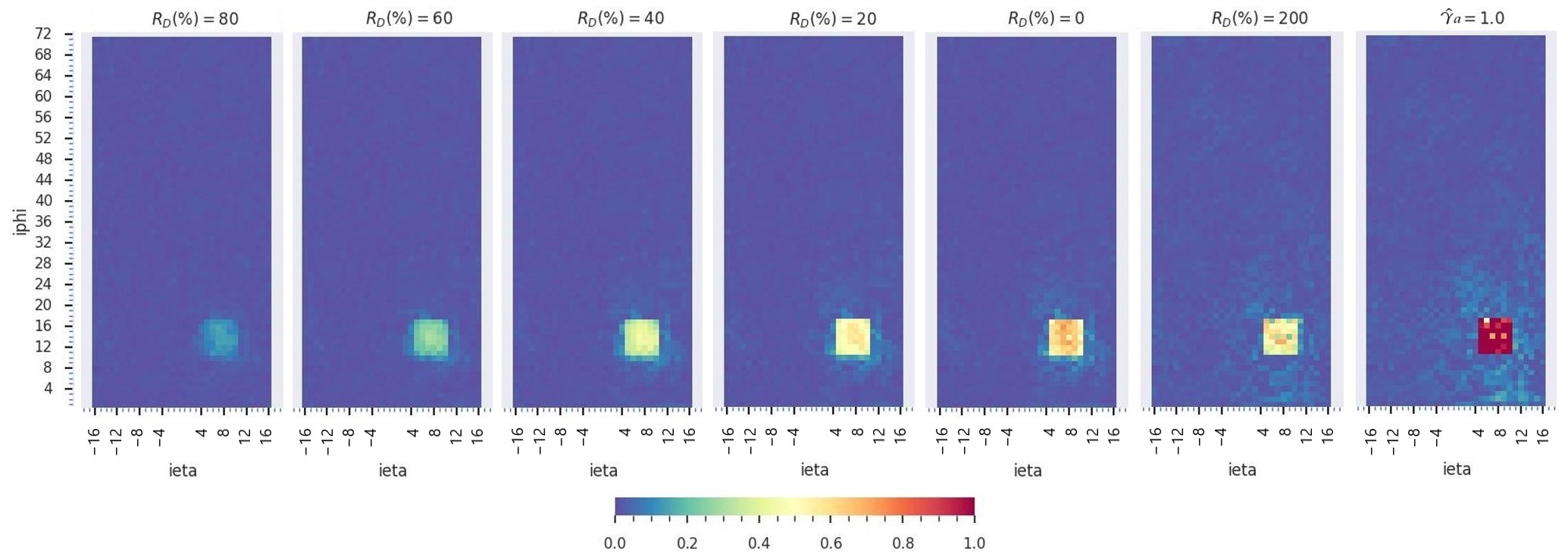}}
% \medskip
\end{adjustwidth}
\caption{{Spatial} %MDPI: 1. Please change the hyphen (-) into a minus sign (−, “U+2212”) in the figure, e.g., “-1” should be “−1”. 2. We moved Figures 10--14 after their first citation. Please confirm.
%Author: we have updated the hyphen accordingly 
 AD $e_{i, MAE}$ on a sample digi-occupancy map at \emph{depth}~$=1$ with degraded \mbox{($0<R_D<100\%$)}, dead ($R_D=0\%$), noisy hot ($R_D=200\%$), and~fully hot ($\hat{\gamma}_a=1.0$) anomaly types: (\textbf{a}) renormalized digi-occupancy map with simulated anomaly channels; the reconstruction error maps of the (\textbf{b}) $\mathcal{M}_{\text{N}}$ model and~(\textbf{c}) the $\mathcal{M}_{\text{T}}$ model. The~anomaly region is localized well with proportional strength to the severity of the anomaly in both models. The~$\mathcal{M}_{\text{T}}$ model has better localization with relatively less dispersion in its anomaly score map. \label{fig:pred_err_window_spatial_idx_0_rd_compare}
}
\end{figure}

Table~\ref{tbl:clf_per_all_anml_channel_tw} presents the AD accuracy of the models on the dead, degrading, fully hot, and~noisy hot channel abnormalities. Both models perform well in the \textit{area under the receiver operating characteristic curve} (AUC) and \textit{false positive rate} (FPR). 
The TL model significantly improves dead and fully hot channel detection but performs slightly lower for the noisy hot channels.
Figures~\ref{fig:pred_err_window_spatial_idx_0_rd_compare} demonstrate the ability of the models to detect and localize the different anomaly types that have been injected at sample channels located at $\{4 < i\eta < 11,~11< i\phi < 19$~\emph{depth}$\}$; the $\mathcal{M}_{\text{T}}$ model accomplishes better detection on the fully hot channels with less dispersion in its anomaly score~values.

\begin{table}[H]\scriptsize
\caption{{AD} %MDPI: Please confirm if all the bold in the table and the table footer should be kept.
%Authors: we removed some of the unnecessary old and kept the remaining.
 performance DQM abnormal~channels.\label{tbl:clf_per_all_anml_channel_tw}}
%\resizebox{0.9\textwidth}{!}{
\begin{tabularx}{\textwidth}{cCCCC}
\toprule
\textbf{Channel Anomaly Type} & \textbf{FPR (90\%) $\mathbold{\downarrow}$} & \textbf{FPR (95\%) $\mathbold{\downarrow}$} & \textbf{FPR (99\%) $\mathbold{\downarrow}$} &\textbf{AUC $\mathbold{\uparrow}$ } \\  \midrule  
\multicolumn{5}{c}{{${\mathcal{M}_{N}}$: ${\mathcal{T}_\text{init}}$ = $\text{RANDOM}$~\text{and}~$\mathcal{T}_\text{train}$ = \text{No-TL}}} \\  \midrule
 Degraded ($R_D = 80\%$)         & $6.281 \times 10^{-4}$                               & $1.519 \times 10^{-3}$                               & $8.741 \times 10^{-3}$                               & 0.993                                               \\ 
                            Degraded ($R_D = 60\%$)        & $5.991 \times 10^{-5}$                               & $1.438 \times 10^{-4}$                               & $8.242 \times 10^{-4}$                               & \textbf{1.000}                   \\  
                            Degraded ($R_D = 40\%$)        & $4.881 \times 10^{-5}$                               & $5.628 \times 10^{-5}$                               & $1.466 \times 10^{-4}$                               & \textbf{1.000}                   \\  
                            Degraded ($R_D = 20\%$)        & $5.870 \times 10^{-5}$                               & $6.273 \times 10^{-5}$                               & $7.342 \times 10^{-5}$                               & \textbf{1.000}                   \\  
                        Dead ($R_D = 0\%$)                     & $6.636 \times 10^{-5}$                               & $7.080 \times 10^{-5}$                               & $8.331 \times 10^{-5}$                               & \textbf{1.000}                   \\  
        Noisy hot ($R_D = 200\%$)                      & $\mathbf{3.300 \times 10^{-4}}$                      & $\mathbf{5.732 \times 10^{-4}}$                      & $\mathbf{1.765 \times 10^{-3}}$                      & \textbf{1.000}                   \\   
        Fully hot ($\gamma _a = \xi, \gamma _a > \gamma _h$)                      & $1.220 \times 10^{-4}$                      & $1.317 \times 10^{-4}$                      & $1.606 \times 10^{-4}$                      & \textbf{1.000}                   \\  \midrule

\multicolumn{5}{c}{{$\mathcal{M}_{T}$: ${\mathcal{T}_\text{init}}$ = \text{TL-7}~ \text{and}~$\mathcal{T}_\text{train}$ = \text{TL-6}}} \\  \midrule
                             Degraded ($R_D = 80\%$)        & $\mathbf{5.019 \times 10^{-4}}$                      & $\mathbf{1.320 \times 10^{-3}}$                      & $\mathbf{6.527 \times 10^{-3}}$                      & \textbf{0.996}                   \\ 
                            Degraded ($R_D = 60\%$)        & $\mathbf{2.118 \times 10^{-5}}$                      & $\mathbf{9.642 \times 10^{-5}}$                      & $\mathbf{8.141 \times 10^{-4}}$                      & \textbf{1.000}                   \\ 
                            Degraded ($R_D = 40\%$)        & $\mathbf{1.614 \times 10^{-6}}$                      & $\mathbf{4.034 \times 10^{-6}}$                      & $\mathbf{7.161 \times 10^{-5}}$                      & \textbf{1.000}                   \\ 
                            Degraded ($R_D = 20\%$)        & $\mathbf{1.614 \times 10^{-6}}$                      & $\mathbf{3.833 \times 10^{-6}}$                      & $\mathbf{8.472 \times 10^{-6}}$                      & \textbf{1.000}        \\   
                        Dead ($R_D = 0\%$)                     & $\mathbf{1.815 \times 10^{-6}}$                      & $\mathbf{4.236 \times 10^{-6}}$                      & $\mathbf{8.472 \times 10^{-6}}$                      & \textbf{1.000}                   \\      
                Noisy hot ($R_D = 200\%$)                      & $1.380 \times 10^{-3}$                               & $2.143 \times 10^{-3}$                               & $4.099 \times 10^{-3}$                               & \textbf{1.000}      \\    
                Fully hot ($\gamma _a = \xi, \gamma _a > \gamma _h$)                      & \textbf{0.000}                               & $\mathbf{6.051 \times 10^{-7}}$                               & $\mathbf{3.631 \times 10^{-5}}$                               & \textbf{1.000}      \\ 
                
\bottomrule
\end{tabularx}
%}
\noindent{\footnotesize{{AUC:} %MDPI: We merged the table footer into one paragraph. Please confirm.
%Authors: we confirm
 \textit{area under the receiver operating characteristic curve} and FPR: \textit{false positive rate}. 
The FPR ($\rho\%$) denotes the FPR score of capturing the $\rho\%$ of the anomalies.
The \textbf{{bold font}} is the better score between {$\mathbf{\mathcal{M}_{\text{N}}}$} and {$\mathbf{\mathcal{M}_{\text{T}}}$}. 
The down arrows ($\downarrow$) indicates that lower is better, and vice~versa for th up arrows ($\uparrow$).}}
\end{table}
\unskip

\begin{figure}[H]

%\captionsetup[subfigure]{justification=centering}
\begin{adjustwidth}{-\extralength}{0cm}\centering
\subfloat[\centering]{\includegraphics[width=1.3\textwidth]{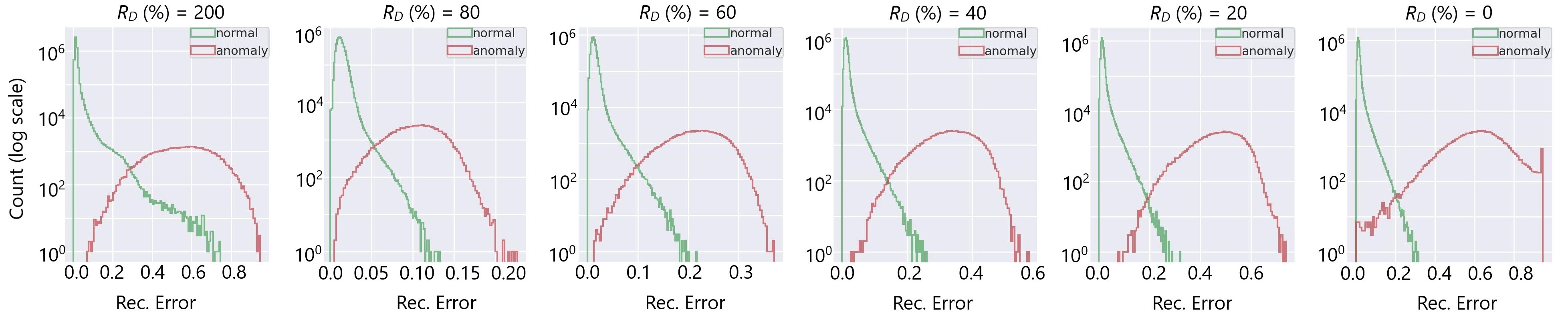}}
% \medskip
 \\
\subfloat[\centering]{\includegraphics[width=1.3\textwidth]{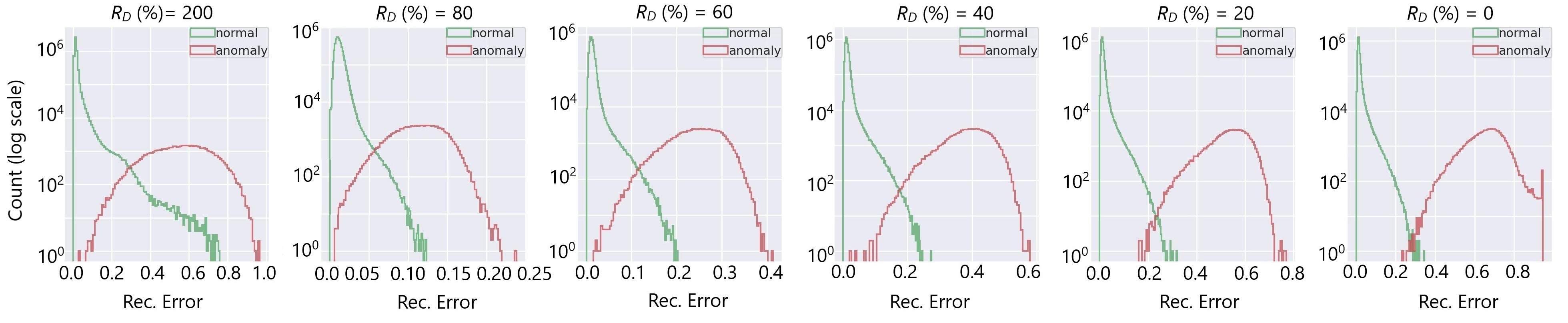}}
% \medskip
\end{adjustwidth}
\caption{AD reconstruction $e_{i, MAE}$ distribution of healthy and anomalous channels at different degradation rates of the simulated anomalies. The~models are (\textbf{a)} $\mathcal{M}_{\text{N}}$ and (\textbf{b}) $\mathcal{M}_{\text{T}}$. The~overlap region decreases substantially as the channel deterioration increases for $R_D<100\%$. However, the~overlap increases for $R_D=200\%$, as~the error increases for the normal channels due to the correlation to adjacent anomalies (as shown in Figure~\ref{fig:rec_error_score_anml_2_tsne}).\label{fig:degrade_hot_to_death_channel__rec_err__pdf_window_spatial_cleaned}
}
\end{figure}

\begin{figure}[H]
\centering
%\captionsetup[subfigure]{justification=centering}
\includegraphics[width=1\textwidth, scale=1]{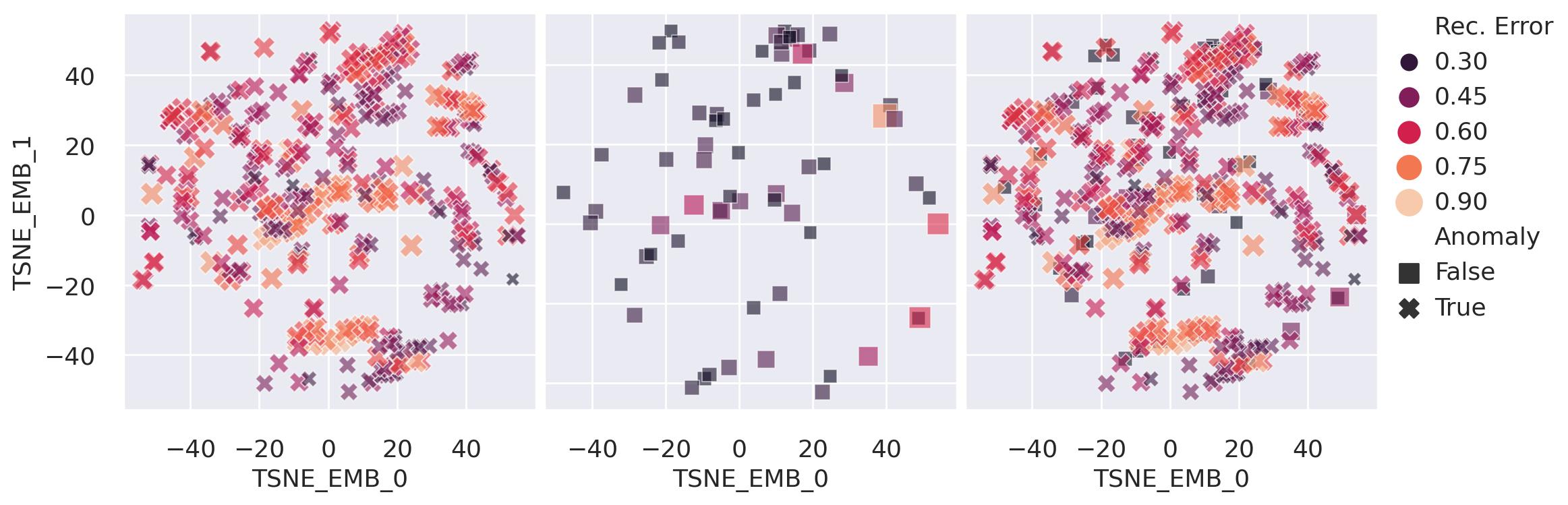}
\caption{Spatio-temporal location embedding for channels with high $e_{i, MAE}$ in the presence of noisy hot anomalies ($R_D=200\%$) with $\mathcal{M}_{\text{T}}$: (left to right) location embedding for the anomaly channels ($Anomaly=True$), the~normal channels ($Anomaly=False$), and both, respectively. We applied t-SNE embedding~\cite{van2008visualizing} to the channels' locations (coordinates: LS, $i\eta$, $i\phi$, and~\emph{depth}) to generate the 2D representation. The~normal channels ($Anomaly=False$) with high reconstruction error occur near the anomalous~channels.\label{fig:rec_error_score_anml_2_tsne}}
\end{figure}

\begin{figure}[H]
%\centering
%\captionsetup[subfigure]{justification=centering}
\includegraphics[width=0.99\textwidth, scale=1]{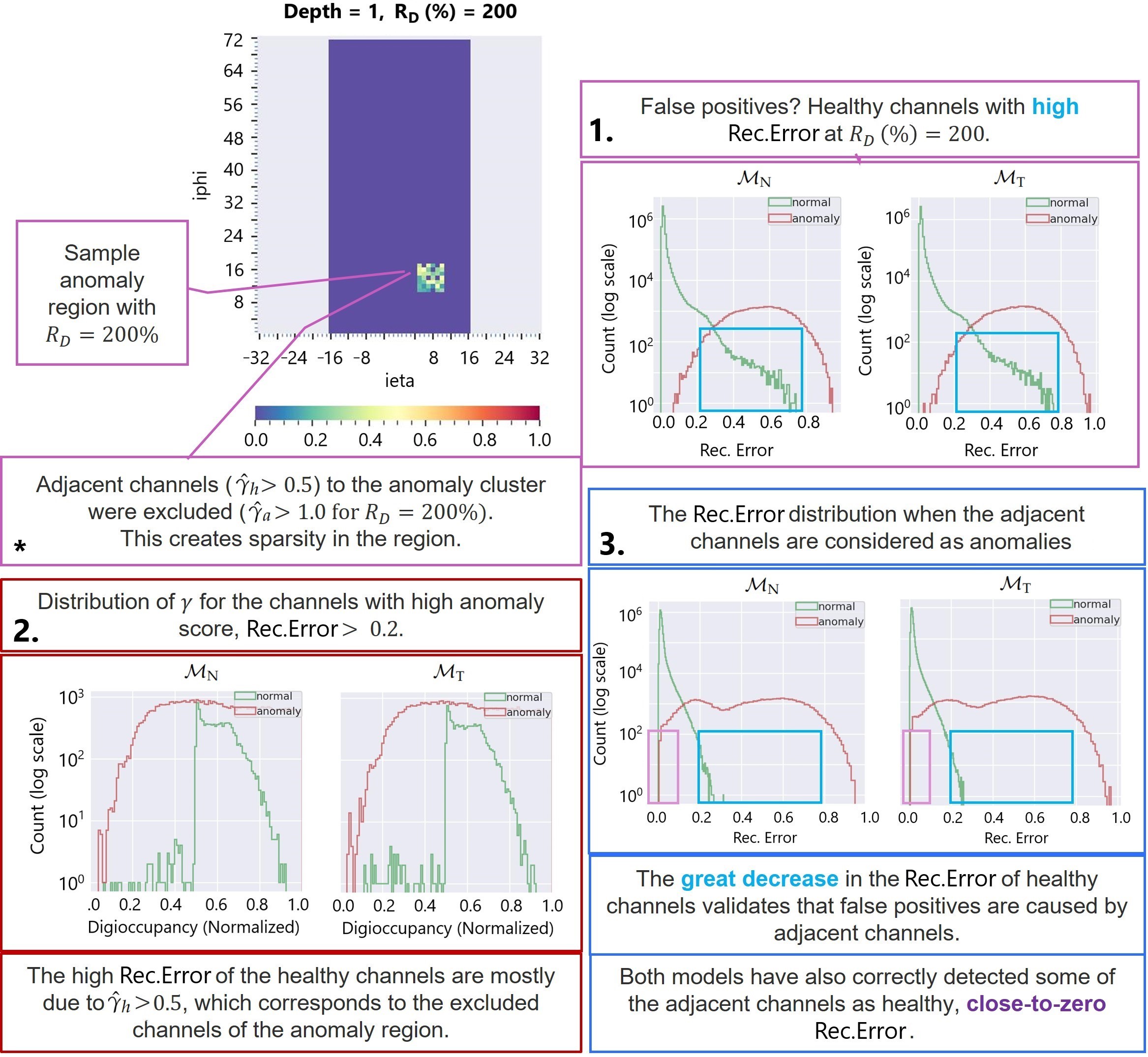}
\caption{{Proximity} %MDPI: 1. Please change the hyphen (-) into a minus sign (−, “U+2212”) in the figure, e.g., “-1” should be “−1”. 2. Please complete the number “0.0” in the left-bottom image 2.
%Author: we have updated the hyphen accordingly 
 effect explanation for false positives in {noisy hot} channel anomaly ($R_D = 200\%$) detection. The~healthy channels with higher AD scores ($e_{i, MAE}>0.2$) belong to the filtered out channels from the anomaly injection $R_D \gamma_{h} > \xi$ (see Equation~\eqref{eq:anml_type_gen}) and generate a high score due to their proximity to the abnormal channels. * Example of anomaly region selection where channels with $R_D \gamma_{h} > \xi$ are excluded from the selection to meet the requirement of $\gamma_{a} \in [0, \xi ]$.\label{fig:stad_tl_dqm__noisy_anomaly_high_false_positive_rate_explain}
}
\end{figure}

Figure~\ref{fig:degrade_hot_to_death_channel__rec_err__pdf_window_spatial_cleaned} portrays the distribution of the AD reconstruction error score ($e_{i, MAE}$) and the overlap region between the healthy and faulty channels at the different degradation rates.
We observed an increase in the error score of the healthy channels for the noisy hot channel anomalies at $R_D=200\%$. 
Close investigation reveals that the healthy channels have higher anomaly scores due to their proximity to the abnormal channels (as illustrated in Figure~\ref{fig:rec_error_score_anml_2_tsne}); the channels are filtered out from the anomaly simulation generation in Equation~\eqref{eq:anml_type_gen} due to $\gamma_{a} = R_D \gamma_{h} > \xi$. Since channels positioned in proximity to the HB segmentation may share common RBX and are exposed to similar collision particles, the~AD AE exploits the correlation for spatial data reconstruction. 
Figure~\ref{fig:stad_tl_dqm__noisy_anomaly_high_false_positive_rate_explain} elaborates further proof of the proximity, showing that the false positive healthy channels with higher anomaly scores ($e_{i, MAE}>0.2$) belong to $R_D \gamma_{h} > \xi$. 

\begin{figure}[H]
\subfloat[\centering]{\includegraphics[width=0.45\textwidth]{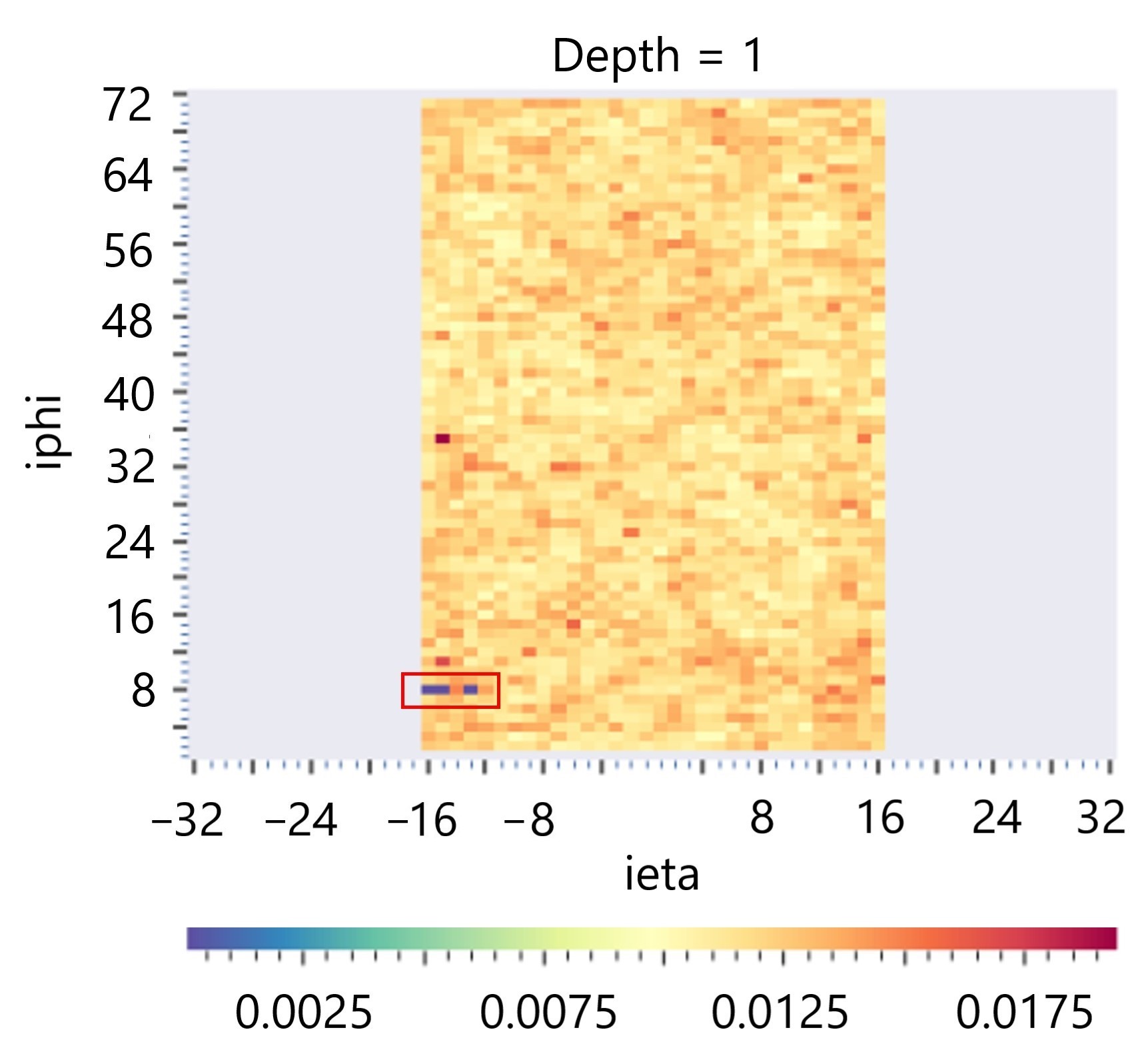}}
%\hfill
\subfloat[\centering]{\includegraphics[width=0.45\textwidth]{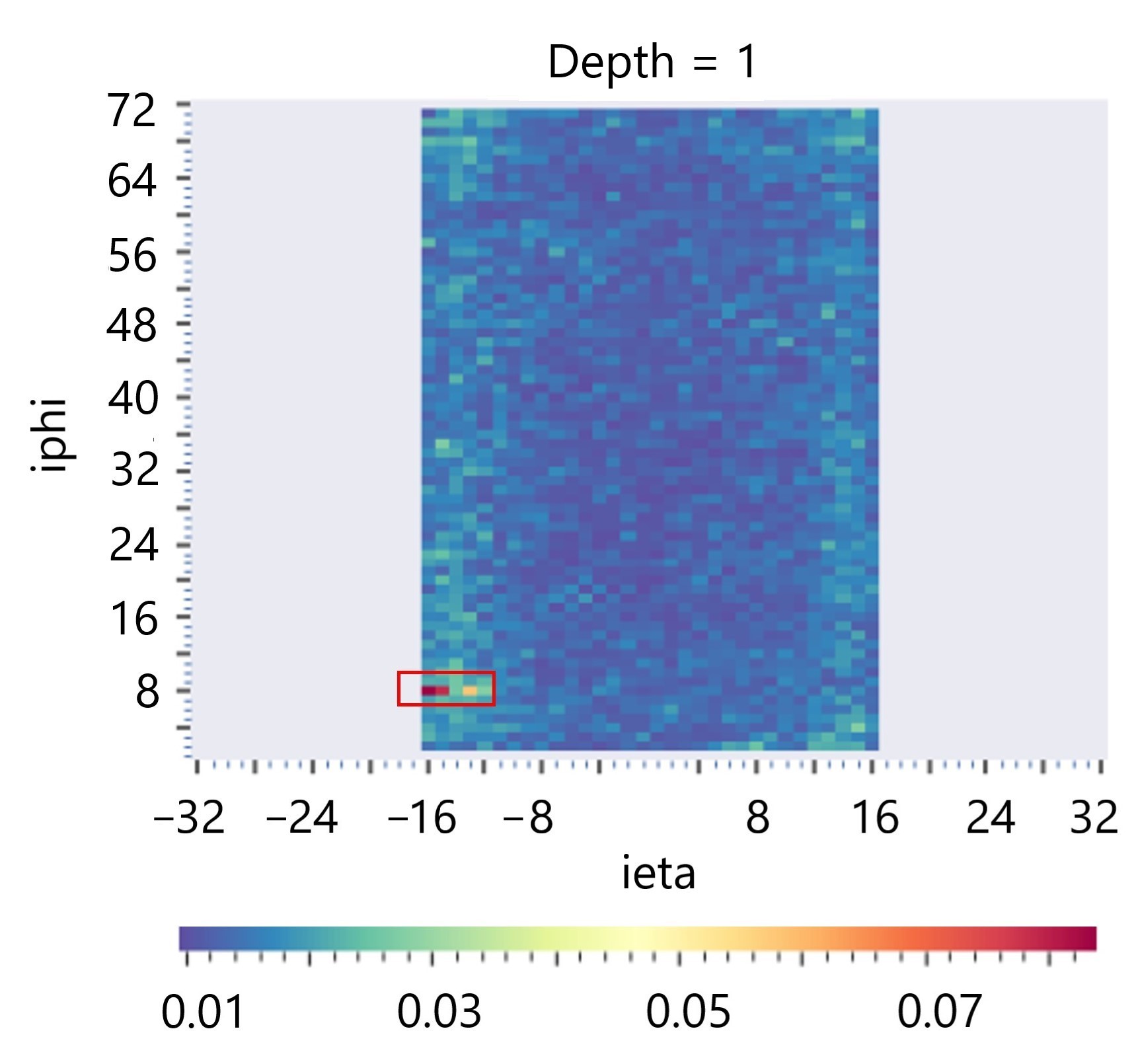}}
%\centering
%\captionsetup[subfigure]{justification=centering}
%\begin{subfigure}[]{0.45\textwidth}
%\centering
%\includegraphics[width=1\textwidth, scale=1]{images/stad_tl_dqm__rec_map_depth_1_without_tl_annot_2.jpg}
%\caption{}
%\medskip
%\end{subfigure}
%\begin{subfigure}[]{0.45\textwidth}
%\centering
%\includegraphics[width=1\textwidth]{images/stad_tl_dqm__rec_map_depth_1_with_tl_annot_2.jpg}
%\caption{}
%\medskip
%\end{subfigure}
\caption{{Spatial} %MDPI: Please change the hyphen (-) into a minus sign (−, “U+2212”) in the figure, e.g., “-1” should be “−1”.
%Author: we have updated the hyphen accordingly 
 AD $e_{i, MAE}$ per channel map at \emph{depth}~$=1$ averaged over the training dataset for the (\textbf{a}) $\mathcal{M}_{\text{N}}$ model, and~(\textbf{b}) $\mathcal{M}_{\text{T}}$ model.  
$\mathcal{M}_{\text{N}}$ reconstructs the real dead channels (in red boxes) as normal, with~very low error scores. In~contrast,~$\mathcal{M}_{\text{T}}$ produces a high error, which signifies detecting the channels as anomalies.}
\label{fig:train_rec_err_mean_map}
\end{figure}

The degrading and dead channels are another major difference between the $\mathcal{M}_{\text{N}}$ and $\mathcal{M}_{\text{T}}$ models. 
Table~\ref{tbl:clf_per_all_anml_channel_tw} shows that the AD slightly deteriorates for the dead channels ($R_D=0\%$) compared to the degraded channel at $R_D=20\%$, defying the expectation of getting better AD on stronger anomalies. 
The error score of the $\mathcal{M}_{\text{N}}$ model drops to zero for dead channels, although the channels have higher error scores at $R_D=60\%$ (see 
Figure~\ref{fig:degrade_hot_to_death_channel__rec_err__pdf_window_spatial_cleaned}a); this is influenced by the presence of real dead channels in the training dataset at the location of $\{i\eta \in [-16, -15, -13], i\phi=8,$ \emph{depth}~$=1\}$ (see Figure~\ref{fig:train_rec_err_mean_map}). Figure~\ref{fig:train_rec_err_mean_map} depicts that the $\mathcal{M}_{\text{N}}$ model learned to reconstruct the real dead channels as healthy (very low error score), whereas $\mathcal{M}_{\text{T}}$ provides a high error, signifying it is detecting the channels as anomalies. 
The results demonstrate TL robustness when a semi-supervised model's training dataset is contaminated with real~anomalies.

\unskip

\unskip

\unskip

\unskip

\section{Conclusions}
\label{sec:conclusion}

We have presented transfer learning on anomaly detection models in the context of high-dimensional spatio-temporal autoencoders.  
We have discussed TL using semi-supervised AD models designed to monitor the Hadron Calorimeter using three-dimensional digi-occupancy maps of the Data Quality Monitoring system. 
We have successfully transferred the AD model, employing convolutional, graph, and~recurrent neural networks, from~the source HCAL Endcap to the target HCAL Barrel calorimeter using several ST TL configurations. 
This study has provided insights into several TL scenarios at the model initialization and training phases in both the encoder and decoder networks. 
Applying TL to the feature extraction networks of the encoder and inner reconstruction networks of the decoder provided promising results in the ST AD. 
The approach has also demonstrated potential leverage for small training datasets, a~significant reduction in training computation, and~an enhancement in robustness against data contamination in the training dataset.
In addition to the similarity between the source and target datasets, the~choice of model settings, such as the target network layers, the~number of training iterations, the~learning rate schedule, and~the temporal state preservation during inference, can influence the performance of TL.
Our study remains relevant to applications in other domains, as~it provides essential understanding of TL on widely utilized hybrid neural network layers applied to ST AD datasets.\vspace{6pt}

%%%%%%%%%%%%%%%%%%%%%%%%%%%%%%%%%%%%%%%%%%
\supplementary{The following supporting information can be downloaded at: \linksupplementary{s1}, {Membership of the CMS-HCAL Collaboration.}}
\authorcontributions{{Conceptualization,} M.W.A., L.W., and D.Y.; data curation, M.W.A., L.W., and D.Y.; formal analysis, M.W.A.; investigation, M.W.A., L.W., P.P., and J.D.; methodology, M.W.A. and C.W.O.; {resources and collision experiment}, L.W., D.Y., P.P., J.D., and the CMS-HCAL Collaboration; software, M.W.A.; supervision, C.W.O.; validation, M.W.A., C.W.O., L.W., and D.Y.; visualization, M.W.A.; writing---original draft, M.W.A.; writing---review and editing, M.W.A., C.W.O., L.W., P.P., J.D, and {the CMS-HCAL Collaboration}. 
All authors have read and agreed to the published version of the~manuscript.}

\funding{{This research received no external funding}
%MDPI: Please add: ``This research received no external funding'' or ``This research was funded by NAME OF FUNDER grant number XXX.'' and  and ``The APC was funded by XXX''. Check carefully that the details given are accurate and use the standard spelling of funding agency names at \url{https://search.crossref.org/funding}, any errors may affect your future funding.
}

\institutionalreview{{Not applicable}
%MDPI: In this section, you should add the Institutional Review Board Statement and approval number, if relevant to your study. You might choose to exclude this statement if the study did not require ethical approval. Please note that the Editorial Office might ask you for further information. Please add “The study was conducted in accordance with the Declaration of Helsinki, and approved by the Institutional Review Board (or Ethics Committee) of NAME OF INSTITUTE (protocol code XXX and date of approval).” for studies involving humans. OR “The animal study protocol was approved by the Institutional Review Board (or Ethics Committee) of NAME OF INSTITUTE (protocol code XXX and date of approval).” for studies involving animals. OR “Ethical review and approval were waived for this study due to REASON (please provide a detailed justification).” OR “Not applicable” for studies not involving humans or animals.
}

\informedconsent{{Not applicable}
%MDPI: Any research article describing a study involving humans should contain this statement. Please add ``Informed consent was obtained from all subjects involved in the study.'' OR ``Patient consent was waived due to REASON (please provide a detailed justification).'' OR ``Not applicable'' for studies not involving humans. You might also choose to exclude this statement if the study did not involve humans. Written informed consent for publication must be obtained from participating patients who can be identified (including by the patients themselves). Please state ``Written informed consent has been obtained from the patient(s) to publish this paper'' if applicable.
}

\dataavailability{{Not applicable}
%MDPI: This section is necessary, please do not delete it. Also, please do not add “Not applicable”. Please note that "Not applicable" is only used for review papers or articles for which no new data were created. Normally, the DAS for research articles should state "Data are contained within the article." or "Data are contained within the article and supplementary materials." Please refer to the complete guideline at https://www.mdpi.com/ethics#_bookmark21.
} 

\acknowledgments{% \section*{Acknowledgments}
We sincerely appreciate the CMS collaboration---specifically the HCAL data performance group, the~HCAL operation group, the~CMS data quality monitoring groups, and~the CMS machine learning core teams. Their technical expertise, diligent follow-up on our work, and~thorough manuscript review have been invaluable. We also thank the collaborators for building and maintaining the detector systems used in our study. We extend our appreciation to CERN for the operations of the LHC accelerator. {The teams at CERN have also received support from the Belgian Fonds de la Recherche Scientifique and Fonds voor Wetenschappelijk Onderzoek; the Brazilian Funding Agencies (CNPq, CAPES, FAPERJ, FAPERGS, and~FAPESP); SRNSF (Georgia); the Bundesministerium f\"ur Bildung und Forschung, the~Deutsche Forschungsgemeinschaft (DFG), under~Germany's Excellence Strategy---EXC 2121 ''Quantum Universe''---390833306, and~under project number 400140256---GRK2497, and~Helmholtz-Gemeinschaft Deutscher Forschungszentren, Germany; the National Research, Development and Innovation Office (NKFIH) (Hungary) under project numbers K~128713, K~143460, and~TKP2021-NKTA-64; the Department of Atomic Energy and the Department of Science and Technology, India; the Ministry of Science, ICT and Future Planning, and~National Research Foundation (NRF), Republic of Korea; the Lithuanian Academy of Sciences; the Scientific and Technical Research Council of Turkey, and~Turkish Energy, Nuclear and Mineral Research Agency; the National Academy of Sciences of Ukraine; the US Department of Energy.} 
%MDPI: Please confirm if these content in Acknowledgments is funding information, if yes, please move it in funding part and remove Acknowledgments part. Please note that this part should not be the same as funding.
%Authors recommend that the acknowledgment remain as it is. It is a CERN requirement to include the listed institutions and organizations in the acknowledgment. The list is authorized and provided by CERN.
}

\conflictsofinterest{{The authors declare no conflict of interest}
%MDPI: Declare conflicts of interest or state ``The authors declare no conflicts of interest.'' Authors must identify and declare any personal circumstances or interest that may be perceived as inappropriately influencing the representation or interpretation of reported research results. Any role of the funders in the design of the study; in the collection, analyses or interpretation of data; in the writing of the manuscript; or in the decision to publish the results must be declared in this section. If there is no role, please state ``The funders had no role in the design of the study; in the collection, analyses, or interpretation of data; in the writing of the manuscript; or in the decision to publish the results''.
} 

% please add the following content in "Author Contributions" section in your manuscript:
% Author Contributions

%%%%%%%%%%%%%%%%%%%%%%%%%%%%%%%%%%%%%%%%%%
\vspace{-3pt}\abbreviations{Abbreviations}{
The following abbreviations are used in this {manuscript:} %MDPI: Please confirm if all the bold in Abbreviations below are unnecesary and can be removed.
%Authors: we removed the bold formatting
\\

\def\NoBoldItem[#1]#2{\item[{\rm #1}]#2}

\vspace{-5pt}\noindent 
\begin{description}[labelwidth=\mylen,nosep]
    \NoBoldItem[AD]           Anomaly Detection  
    \NoBoldItem[AE]           Autoencoder  
    \NoBoldItem[AUC]         Area Under the Curve   
    \NoBoldItem[CMS]          Compact Muon Solenoid  
    \NoBoldItem[CNN]          Convolutional Neural Network  
    \NoBoldItem[DL]          Deep Learning  
    \NoBoldItem[DQM]          Data Quality Monitoring  
    \NoBoldItem[ECAL]       Electromagnetic Calorimeter 
    \NoBoldItem[FC]          Fully Connected Neural Network  
    \NoBoldItem[FPR]        False Positive Rate  
    \NoBoldItem[GNN]          Graph Neural Network  
    \NoBoldItem[\textsc{GraphSTAD}]          Graph-based ST AD model  
    \NoBoldItem[HCAL]          Hadron Calorimeter  
    \NoBoldItem[{HB, HE, HF, HO}]         {HCAL Barrel, HCAL Endcap, HCAL Forward, and~HCAL Outer subdetectors}        
    \NoBoldItem[HPD]        Hybrid Photodiode Transducers 
    \NoBoldItem[KL]          Kullback--Leibler divergence  
    \NoBoldItem[LHC]          Large Hadron Collider  
    \NoBoldItem[LR]         Learning Rate 
    \NoBoldItem[LS]       Luminosity Section 
    \NoBoldItem[LSTM]  Long Short-Term Memory  
    \NoBoldItem[{MAE, MSE}]          {Mean Absolute Error, Mean Squared Error} 
    \NoBoldItem[QIE]         Charge Integrating and Encoding 
    \NoBoldItem[RBX]          Readout Box  
    \NoBoldItem[RNN]          Recurrent Neural Network  
    \NoBoldItem[SiPM]         Silicon Photo Multipliers 
    \NoBoldItem[ST]           Spatio-Temporal 
    \NoBoldItem[TL]       Transfer Learning 
    \NoBoldItem[{${\gamma}$, ${\hat{\gamma}}$, ${\gamma_h}$, ${\gamma_a}$}]     {Digi-occupancy map, renomalized $\gamma$ by $\xi$, healthy $\gamma$, anomalous $\gamma$}
    \NoBoldItem[{${\beta}$, ${\xi}$}]       {Received luminosity, number of collision events}
    \NoBoldItem[{${i\eta}$, ${i\phi}$, ${depth}$}]      {$ieta$, $iphi$, and~$\emph{depth}$ axes of the HCAL channels  }
    \NoBoldItem[{$\mathcal{F}_{\theta, \omega}$, $\mathcal{E}_\theta$, $\mathcal{D}_\omega$}]  {Mathematical notation of an AE model, encoder of $\mathcal{F}_{\theta, \omega}$, decoder of $\mathcal{F}_{\theta, \omega}$}
    \NoBoldItem[{$\mathcal{L}_{MSE}$, $e_{i, MAE}$}]       {AE reconstruction MSE loss score, MAE AD score}
    \NoBoldItem[ $R_D$]    Degradation factor of channel anomaly
     \NoBoldItem[{$\mathcal{M}_e$, $\mathcal{M}_b$}]      {AE model of the TL source HE system, AE model of the TL target HB system} 
    \NoBoldItem[{$\mathcal{M}_{\text{N}}$, $\mathcal{M}_{\text{T}}$}]      {Model trained without TL, model trained with TL}
    \NoBoldItem[{$\mathcal{T}_{\text{init}}$, $\mathcal{T}_{\text{train}}$}]   {TL during initialization phase, TL during training phase (frozen parameters)}       
\end{description}
}

% \footnotesize
\begin{adjustwidth}{-\extralength}{0cm}
\reftitle{{References}}
%\bibliography{manuscript.bib}

\begin{thebibliography}{999}

\bibitem[Atluri \em{et~al.}(2018)Atluri, Karpatne, and Kumar]{atluri2018spatio}
Atluri, G.; Karpatne, A.; Kumar, V.
\newblock Spatio-temporal data mining: A survey of problems and methods.
\newblock {\em ACM Comput. Surv.} {\bf 2018}, {\em 51},~{83.} %MDPI: Newly added information. Please confirm all highlights.
%Author: we confirm 

\bibitem[Chang \em{et~al.}(2022)Chang, Tu, Xie, Luo, Zhang, Sui, and Yuan]{chang2022video}
Chang, Y.; Tu, Z.; Xie, W.; Luo, B.; Zhang, S.; Sui, H.; Yuan, J.
\newblock Video anomaly detection with spatio-temporal dissociation.
\newblock {\em Pattern Recognit.} {\bf 2022}, {\em 122},~108213.

\bibitem[Deng \em{et~al.}(2022)Deng, Lian, Huang, and Chen]{deng2022graph}
Deng, L.; Lian, D.; Huang, Z.; Chen, E.
\newblock Graph convolutional adversarial networks for spatiotemporal anomaly detection.
\newblock {\em IEEE Trans. Neural Netw. Learn. Syst.} {\bf 2022}, {\em 33},~2416--2428.

\bibitem[Ti{\v{s}}ljari{\'c} \em{et~al.}(2021)Ti{\v{s}}ljari{\'c}, Fernandes, Cari{\'c}, and Gama]{tivsljaric2021spatiotemporal}
Ti{\v{s}}ljari{\'c}, L.; Fernandes, S.; Cari{\'c}, T.; Gama, J.
\newblock Spatiotemporal road traffic anomaly detection: A tensor-based approach.
\newblock {\em Appl. Sci.} {\bf 2021}, {\em 11},~12017.

\bibitem[Fathizadan \em{et~al.}(2023)Fathizadan, Ju, Lu, and Yang]{fathizadan2023deep}
Fathizadan, S.; Ju, F.; Lu, Y.; Yang, Z.
\newblock Deep spatio-temporal anomaly detection in laser powder bed fusion.
\newblock {\em IEEE Trans. Autom. Sci. Eng.} {\bf {2023}}, {\em 21},~{5227--5239}.

\bibitem[Asres \em{et~al.}(2023)Asres, Omlin, Wang, Yu, Parygin, Dittmann, Karapostoli, Seidel, Venditti, Lambrecht, et~al.]{mulugeta2022dqm}
Asres, M.W.; Omlin, C.W.; Wang, L.; Yu, D.; Parygin, P.; Dittmann, J.; Karapostoli, G.; Seidel, M.; Venditti, R.; Lambrecht, L.;  et~al.
\newblock Spatio-temporal anomaly detection with graph networks for data quality monitoring of the {Hadron Calorimeter}.
\newblock {\em Sensors} {\bf 2023}, {\em 23},~9679.

\bibitem[Zhao \em{et~al.}(2022)Zhao, Deng, Chen, Guo, Yang, Kieu, Huang, Pedersen, Zheng, and Jensen]{zhao2022comparative}
Zhao, Y.; Deng, L.; Chen, X.; Guo, C.; Yang, B.; Kieu, T.; Huang, F.; Pedersen, T.B.; Zheng, K.; Jensen, C.S.
\newblock A comparative study on unsupervised anomaly detection for time series: Experiments and analysis.
\newblock {\em arXiv} {\bf 2022}, arXiv:2209.04635.

\bibitem[Chalapathy and Chawla(2019)]{chalapathy2019deep}
Chalapathy, R.; Chawla, S.
\newblock Deep learning for anomaly detection: A survey. {\em arXiv} {\bf 2019},
\newblock {arXiv:1901.03407}.

\bibitem[Cook \em{et~al.}(2019)Cook, M{\i}s{\i}rl{\i}, and Fan]{cook2019anomaly}
Cook, A.A.; M{\i}s{\i}rl{\i}, G.; Fan, Z.
\newblock Anomaly detection for {IoT} time-series data: A survey.
\newblock {\em IEEE Internet Things J.} {\bf 2019}, {\em 7},~6481--6494.

\bibitem[Wang and Liu(2024)]{wang2024self}
Wang, Y.; Liu, G.
\newblock Self-supervised dam deformation anomaly detection based on temporal--spatial contrast learning.
\newblock {\em Sensors} {\bf 2024}, {\em 24},~5858.

\bibitem[Yang \em{et~al.}(2025)Yang, Chu, Guo, and Ge]{yang2025weighted}
Yang, J.; Chu, H.; Guo, L.; Ge, X.
\newblock A Weighted-Transfer Domain-Adaptation Network Applied to Unmanned Aerial Vehicle Fault Diagnosis.
\newblock {\em Sensors} {\bf 2025}, {\em 25},~1924.

\bibitem[Wang \em{et~al.}(2021)Wang, Miao, Li, and Cao]{wang2021spatio}
Wang, S.; Miao, H.; Li, J.; Cao, J.
\newblock Spatio-temporal knowledge transfer for urban crowd flow prediction via deep attentive adaptation networks.
\newblock {\em IEEE Trans. Intell. Transp. Syst.} {\bf 2021}, {\em 23},~4695--4705.

\bibitem[Yu \em{et~al.}(2022)Yu, Xiu, and Li]{yu2022survey}
Yu, F.; Xiu, X.; Li, Y.
\newblock A survey on deep transfer learning and beyond.
\newblock {\em Mathematics} {\bf 2022}, {\em 10},~3619.

\bibitem[Shao \em{et~al.}(2018)Shao, McAleer, Yan, and Baldi]{shao2018highly}
Shao, S.; McAleer, S.; Yan, R.; Baldi, P.
\newblock Highly accurate machine fault diagnosis using deep transfer learning.
\newblock {\em IEEE Trans. Ind. Inform.} {\bf 2018}, {\em 15},~2446--2455.

\bibitem[Laptev \em{et~al.}(2018)Laptev, Yu, and Rajagopal]{laptev2018reconstruction}
Laptev, N.; Yu, J.; Rajagopal, R.
\newblock Reconstruction and regression loss for time-series transfer learning.
\newblock In Proceedings of the Special Interest Group on SIGKDD, {London, UK, 19--23 August} 2018; Volume 20.
%Authors: we confirm

\bibitem[Gupta \em{et~al.}(2018)Gupta, Malhotra, Vig, and Shroff]{gupta2018transfer}
Gupta, P.; Malhotra, P.; Vig, L.; Shroff, G.
\newblock Transfer learning for clinical time series analysis using recurrent neural networks.
\newblock {\em arXiv} {\bf 2018}, arXiv:1807.01705.

\bibitem[Boull{\'e} \em{et~al.}(2020)Boull{\'e}, Dallas, Nakatsukasa, and Samaddar]{boulle2020classification}
Boull{\'e}, N.; Dallas, V.; Nakatsukasa, Y.; Samaddar, D.
\newblock Classification of chaotic time series with deep learning.
\newblock {\em Phys. D Nonlinear Phenom.} {\bf 2020}, {\em 403},~132261.

\bibitem[Wang \em{et~al.}(2018)Wang, Geng, Ma, Liu, and Yang]{wang2018cross}
Wang, L.; Geng, X.; Ma, X.; Liu, F.; Yang, Q.
\newblock Cross-city transfer learning for deep spatio-temporal prediction.
\newblock {\em arXiv} {\bf 2018}, arXiv:1802.00386.

\bibitem[Hijazi \em{et~al.}(2023)Hijazi, Dehghanian, and Wang]{hijazi2023transfer}
Hijazi, M.; Dehghanian, P.; Wang, S.
\newblock Transfer learning for transient stability predictions in modern power systems under enduring topological changes.
\newblock {\em IEEE Trans. Autom. Sci. Eng.} {\bf {2023}}.

\bibitem[Shao \em{et~al.}(2014)Shao, Zhu, and Li]{shao2014transfer}
Shao, L.; Zhu, F.; Li, X.
\newblock Transfer learning for visual categorization: A survey.
\newblock {\em IEEE Trans. Neural Netw. Learn. Syst.} {\bf 2014}, {\em 26},~1019--1034.

\bibitem[Niu \em{et~al.}(2020)Niu, Liu, Wang, and Song]{niu2020decade}
Niu, S.; Liu, Y.; Wang, J.; Song, H.
\newblock A decade survey of transfer learning (2010--2020).
\newblock {\em IEEE Trans. Artif. Intell.} {\bf 2020}, {\em 1},~151--166.

\bibitem[Adama \em{et~al.}(2021)Adama, Lotfi, and Ranson]{adama2021survey}
Adama, D.A.; Lotfi, A.; Ranson, R.
\newblock A survey of vision-based transfer learning in human activity recognition.
\newblock {\em Electronics} {\bf 2021}, {\em 10},~2412.

\bibitem[Chato and Regentova(2023)]{chato2023survey}
Chato, L.; Regentova, E.
\newblock Survey of transfer learning approaches in the machine learning of digital health sensing data.
\newblock {\em J. Pers. Med.} {\bf 2023}, {\em 13},~1703.

\bibitem[Russakovsky \em{et~al.}(2015)Russakovsky, Deng, Su, Krause, Satheesh, Ma, Huang, Karpathy, Khosla, Bernstein, Berg, and Fei-Fei]{ILSVRC15}
Russakovsky, O.; Deng, J.; Su, H.; Krause, J.; Satheesh, S.; Ma, S.; Huang, Z.; Karpathy, A.; Khosla, A.; Bernstein, M.;  et~al.
\newblock {ImageNet} Large scale visual recognition challenge.
\newblock {\em Int. J. Comput. Vis.} {\bf 2015}, {\em 115},~211--252.

\bibitem[Devlin \em{et~al.}(2018)Devlin, Chang, Lee, and Toutanova]{devlin2018bert}
Devlin, J.; Chang, M.W.; Lee, K.; Toutanova, K.
\newblock {BERT}: Pre-training of deep bidirectional transformers for language understanding. \emph{arXiv} {\bf 2018},
\newblock {arXiv:1810.04805}.

\bibitem[Sarker \em{et~al.}(2021)Sarker, Losada-Guti{\'e}rrez, Marron-Romera, Fuentes-Jim{\'e}nez, and Luengo-S{\'a}nchez]{sarker2021semi}
Sarker, M.I.; Losada-Guti{\'e}rrez, C.; Marron-Romera, M.; Fuentes-Jim{\'e}nez, D.; Luengo-S{\'a}nchez, S.
\newblock Semi-supervised anomaly detection in video-surveillance scenes in the wild.
\newblock {\em Sensors} {\bf 2021}, {\em 21},~3993.

\bibitem[Natha \em{et~al.}(2025)Natha, Ahmed, Siraj, Lagari, Altamimi, and Chandio]{natha2025deep}
Natha, S.; Ahmed, F.; Siraj, M.; Lagari, M.; Altamimi, M.; Chandio, A.A.
\newblock Deep {BiLSTM} attention model for spatial and temporal anomaly detection in video surveillance.
\newblock {\em Sensors} {\bf 2025}, {\em 25},~251.

\bibitem[Guo \em{et~al.}(2018)Guo, Li, Zheng, Wang, and Yu]{guo2018citytransfer}
Guo, B.; Li, J.; Zheng, V.W.; Wang, Z.; Yu, Z.
\newblock {CityTransfer}: Transferring inter-and intra-city knowledge for chain store site recommendation based on multi-source urban data.
\newblock {\em Proc. ACM Interactive Mobile Wearable Ubiquitous Technol.} {\bf 2018}, {\em 1},~{1--23}.
%corrected

\bibitem[Evans and Bryant(2008)]{evans2008lhc}
Evans, L.; Bryant, P.
\newblock {LHC} machine.
\newblock {\em J. Instrum.} {\bf 2008}, {\em 3},~S08001.

\bibitem[{The CMS Collaboration}(2023)]{cms2023development}
{The CMS Collaboration}.
\newblock Development of the {CMS} detector for the {CERN} {LHC} {Run} 3.
\newblock {\em arXiv} {\bf 2023}, arXiv:2309.05466.

\bibitem[{The CMS Collaboration} \em{et~al.}(2008){The CMS Collaboration}, Chatrchyan, Hmayakyan, Khachatryan, Sirunyan, Adam, Bauer, Bergauer, Bergauer, Dragicevic, et~al.]{collaboration2008cms}
{The CMS Collaboration}; Chatrchyan, S.; Hmayakyan, G.; Khachatryan, V.; Sirunyan, A.; Adam, W.; Bauer, T.; Bergauer, T.; Bergauer, H.; Dragicevic, M.;  et~al.
\newblock The {CMS} experiment at the {CERN} {LHC}.
\newblock {\em J. Instrum.} {\bf 2008}, {\em 3},~S08004.

\bibitem[Azzolini \em{et~al.}(2019)Azzolini, Bugelskis, Hreus, Maeshima, Fernandez, Norkus, Fraser, Rovere, Schneider, et~al.]{azzolini2019data}
Azzolini, V.; Bugelskis, D.; Hreus, T.; Maeshima, K.; Fernandez, M.J.; Norkus, A.; Fraser, P.J.; Rovere, M.; Schneider, M.A.;  et~al.
\newblock The data quality monitoring software for the {CMS} experiment at the {LHC}: Past, present and future.
\newblock \emph{EPJ {Web} Conf.} \textbf{2019}, \linebreak  \emph{214}, 02003.

\bibitem[Tuura \em{et~al.}(2010)Tuura, Meyer, Segoni, and Della~Ricca]{tuura2010cms}
Tuura, L.; Meyer, A.; Segoni, I.; Della~Ricca, G.
\newblock {CMS} data quality monitoring: Systems and experiences.
\newblock \emph{J. Phys. Conf. Ser.} \textbf{2010}, \emph{219}, 072020.

\bibitem[De~Guio and {The CMS Collaboration}(2014)]{de2014cms}
De~Guio, F.; {The CMS Collaboration}.
\newblock The {CMS} data quality monitoring software: Experience and future prospects.
\newblock \emph{J. Phys. Conf. Ser.} \textbf{2014}, \emph{513}, 032024.

\bibitem[Azzolin \em{et~al.}(2019)Azzolin, Andrews, Cerminara, Dev, Jessop, Marinelli, Mudholkar, Pierini, Pol, and Vlimant]{azzolin2019improving}
Azzolin, V.; Andrews, M.; Cerminara, G.; Dev, N.; Jessop, C.; Marinelli, N.; Mudholkar, T.; Pierini, M.; Pol, A.; Vlimant, J.R.
\newblock Improving data quality monitoring via a partnership of technologies and resources between the {CMS} experiment at {CERN} and industry.
\newblock \emph{EPJ {Web} Conf.} \textbf{2019}, \emph{214}, 01007.

\bibitem[Asres \em{et~al.}(2021)Asres, Cummings, Parygin, Khukhunaishvili, Toms, Campbell, Cooper, Yu, Dittmann, and Omlin]{asres2021unsupervised}
Asres, M.W.; Cummings, G.; Parygin, P.; Khukhunaishvili, A.; Toms, M.; Campbell, A.; Cooper, S.I.; Yu, D.; Dittmann, J.; Omlin, C.W.
\newblock Unsupervised deep variational model for multivariate sensor anomaly detection.
\newblock In Proceedings of the International Conference on Progress in Informatics and Computing, {Shanghai, China, 17--19 December} 2021; pp. 364--371.
%Authors: we confirm

\bibitem[Asres \em{et~al.}(2022)Asres, Cummings, Khukhunaishvili, Parygin, Cooper, Yu, Dittmann, and Omlin]{asres2022long}
Asres, M.W.; Cummings, G.; Khukhunaishvili, A.; Parygin, P.; Cooper, S.I.; Yu, D.; Dittmann, J.; Omlin, C.W.
\newblock Long horizon anomaly prediction in multivariate time series with causal autoencoders.
\newblock \emph{Phm Soc. Eur. Conf.} \textbf{2022}, \emph{7}, 21--31.

\bibitem[{The CMS-ECAL Collaboration} \em{et~al.}(2023){The CMS-ECAL Collaboration} et~al.]{cms2023autoencoder}
{The CMS-ECAL Collaboration}.
\newblock Autoencoder-based Anomaly Detection System for Online Data Quality Monitoring of the {CMS} Electromagnetic Calorimeter.
\newblock {\em arXiv} {\bf 2023}, arXiv:2309.10157.

\bibitem[Pol \em{et~al.}(2019{\natexlab{a}})Pol, Azzolini, Cerminara, De~Guio, Franzoni, Pierini, Sirok{\`y}, and Vlimant]{pol2019anomaly}
Pol, A.A.; Azzolini, V.; Cerminara, G.; De~Guio, F.; Franzoni, G.; Pierini, M.; Sirok{\`y}, F.; Vlimant, J.R.
\newblock Anomaly detection using deep autoencoders for the assessment of the quality of the data acquired by the {CMS} experiment.
\newblock \emph{EPJ {Web} Conf.} \textbf{2019}, \emph{214}, 06008.

\bibitem[Pol \em{et~al.}(2019{\natexlab{b}})Pol, Cerminara, Germain, Pierini, and Seth]{pol2019detector}
Pol, A.A.; Cerminara, G.; Germain, C.; Pierini, M.; Seth, A.
\newblock Detector monitoring with artificial neural networks at the {CMS} experiment at the {CERN} {Large Hadron Collider}.
\newblock {\em Comput. Softw. Big Sci.} {\bf 2019}, {\em 3},~3.

\bibitem[Parra \em{et~al.}(2024)Parra, Pardi{\~n}as, P{\'e}rez, Janisch, Klaver, Leh{\'e}ricy, and Serra]{parra2024human}
Parra, O.J.; Pardi{\~n}as, J.G.; P{\'e}rez, L.D.P.; Janisch, M.; Klaver, S.; Leh{\'e}ricy, T.; Serra, N.
\newblock Human-in-the-loop reinforcement learning for data quality monitoring in particle physics experiments.
\newblock {\em arXiv} {\bf 2024}, arXiv:2405.15508.

\bibitem[Xie \em{et~al.}(2020)Xie, Li, Liu, Du, Yang, and Zhang]{xie2020urban}
Xie, P.; Li, T.; Liu, J.; Du, S.; Yang, X.; Zhang, J.
\newblock Urban flow prediction from spatiotemporal data using machine learning: A survey.
\newblock {\em Inf. Fusion} {\bf 2020}, {\em 59},~1--12.

\bibitem[Lai \em{et~al.}(2025)Lai, Zhu, Li, Lan, and Zuo]{lai2025stglr}
Lai, Y.; Zhu, Y.; Li, L.; Lan, Q.; Zuo, Y.
\newblock {STGLR}: A spacecraft anomaly detection method based on spatio-temporal graph learning.
\newblock {\em Sensors} {\bf 2025}, {\em 25},~310.

\bibitem[Strobbe(2017)]{strobbe2017upgrade}
Strobbe, N.
\newblock The upgrade of the {CMS Hadron Calorimeter} with {Silicon} photomultipliers.
\newblock {\em J. Instrum.} {\bf 2017}, {\em 12},~C01080.

\bibitem[Huber \em{et~al.}(2024)Huber, Inderka, and Steinhage]{huber2024leveraging}
Huber, F.; Inderka, A.; Steinhage, V.
\newblock Leveraging remote sensing data for yield prediction with deep transfer learning.
\newblock {\em Sensors} {\bf 2024}, {\em 24},~770.

\bibitem[Wu \em{et~al.}(2020)Wu, Liu, Li, Sun, and Shen]{wu2020fast}
Wu, P.; Liu, J.; Li, M.; Sun, Y.; Shen, F.
\newblock Fast sparse coding networks for anomaly detection in videos.
\newblock {\em Pattern Recognit.} {\bf 2020}, {\em 107},~107515.

\bibitem[Hasan \em{et~al.}(2016)Hasan, Choi, Neumann, Roy-Chowdhury, and Davis]{hasan2016learning}
Hasan, M.; Choi, J.; Neumann, J.; Roy-Chowdhury, A.K.; Davis, L.S.
\newblock Learning temporal regularity in video sequences.
\newblock In Proceedings of the Computer Vision and Pattern Recognition, {Las Vegas, NV, USA, 27--30 June} 2016; pp. 733--742.
%Authors: we confirm

\bibitem[Luo \em{et~al.}(2019)Luo, Liu, Lian, Tang, Duan, Peng, and Gao]{luo2019video}
Luo, W.; Liu, W.; Lian, D.; Tang, J.; Duan, L.; Peng, X.; Gao, S.
\newblock Video anomaly detection with sparse coding inspired deep neural networks.
\newblock {\em IEEE Trans. Pattern Anal. Mach. Intell.} {\bf 2019}, {\em 43},~1070--1084.

\bibitem[Hsu(2017)]{hsu2017anomaly}
Hsu, D.
\newblock Anomaly detection on graph time series.
\newblock {\em arXiv} {\bf 2017}, arXiv:1708.02975.

\bibitem[Focardi(2012)]{focardi2012status}
Focardi, E.
\newblock Status of the {CMS} detector.
\newblock {\em Phys. Procedia} {\bf 2012}, {\em 37},~119--127.

\bibitem[Neutelings(2023)]{tikz2023}
Neutelings, I.
\newblock {CMS} Coordinate System. 2023.  Available {online:} 
 \url{https://tikz.net/axis3d_cms/}
\newblock (accessed on 14 December 2023).
%Author: we confirm

\bibitem[Cheung \em{et~al.}(2012)Cheung, {The CMS collaboration}, et~al.]{cheung2012cms}
Cheung, H.W.; {The CMS Collaboration}.
\newblock {CMS}: Present status, limitations, and upgrade plans.
\newblock {\em Phys. Procedia} {\bf 2012}, {\em 37},~128--137.

\bibitem[Virdee and {The CMS Collaboration}(1999)]{collaboration1999cms}
Virdee, T.; {The CMS Collaboration}.
\newblock The {CMS} experiment at the {CERN} {LHC}.
\newblock In Proceedings of the 6th International Symposium on Particles, Strings and Cosmology, {Boston, MA, USA, 22--29 March} {1999}.

\bibitem[Wen and Keyes(2019)]{wen2019time}
Wen, T.; Keyes, R.
\newblock Time series anomaly detection using convolutional neural networks and transfer learning.
\newblock {\em arXiv} {\bf 2019}, arXiv:1905.13628.

\bibitem[Kingma and Welling(2013)]{kingma2013auto}
Kingma, D.P.; Welling, M.
\newblock Auto-encoding variational bayes.
\newblock {\em arXiv} {\bf 2013}, arXiv:1312.6114.

\bibitem[Van~Laarhoven(2017)]{van2017l2}
Van~Laarhoven, T.
\newblock L2 regularization versus batch and weight normalization.
\newblock {\em arXiv} {\bf 2017}, arXiv:1706.05350.

\bibitem[Kingma and Ba(2014)]{kingma2014adam}
Kingma, D.P.; Ba, J.
\newblock Adam: A method for stochastic optimization.
\newblock {\em arXiv} {\bf 2014}, arXiv:1412.6980.

\bibitem[He \em{et~al.}(2015)He, Zhang, Ren, and Sun]{he2015delving}
He, K.; Zhang, X.; Ren, S.; Sun, J.
\newblock Delving deep into rectifiers: Surpassing human-level performance on imagenet classification.
\newblock In Proceedings of the International Conference on Computer Vision, {Santiago, Chile, 13--16 December} 2015; pp. 1026--1034.
%Author: we confirm

\bibitem[Smith and Topin(2019)]{smith2019super}
Smith, L.N.; Topin, N.
\newblock Super-convergence: Very fast training of neural networks using large learning rates.
\newblock In Proceedings of the {Artificial Intelligence and Machine Learning for Multi-Domain Operations Applications, Baltimore, MD, USA, 14--18 April} 2019; Volume 11006, pp. 369--386.

\bibitem[Van~der Maaten and Hinton(2008)]{van2008visualizing}
Van~der Maaten, L.; Hinton, G.
\newblock Visualizing data using {t-SNE}.
\newblock {\em JMLR} {\bf 2008}, {\em 9}, {2579--2605}.

\end{thebibliography}

\PublishersNote{}
\end{adjustwidth}

\end{document}